\documentclass{article}

    \usepackage[final]{neurips_2024}

\usepackage[utf8]{inputenc} % allow utf-8 input
\usepackage[T1]{fontenc}    % use 8-bit T1 fonts
\usepackage{url}            % simple URL typesetting
\usepackage{booktabs}       % professional-quality tables
\usepackage{amsfonts}       % blackboard math symbols
\usepackage{nicefrac}       % compact symbols for 1/2, etc.
\usepackage{microtype}      % microtypography
\usepackage{xcolor}         % colors
\usepackage{xspace}
\usepackage{graphicx}
\usepackage{subfigure}
\usepackage{booktabs} % for professional tables
\usepackage{multirow}
\usepackage{adjustbox}
\usepackage{enumitem}
\usepackage{xspace}

\definecolor{darker}{rgb}{0,0.15,0.7}
\usepackage[colorlinks, urlcolor=darker, citecolor=darker, linkcolor=darker]{hyperref} 

\definecolor{mygreen}{RGB}{65, 183, 41}

\newcommand{\method}{{ChatQA}\xspace}

\title{\method: Surpassing GPT-4 on Conversational QA and RAG}

\author{%
  Zihan Liu~\thanks{Correspondence to: Zihan Liu <zihanl@nvidia.com>, Wei Ping <wping@nvidia.com>} ~$^{1}$
  \And
  Wei Ping~$^{*}$~$^{1}$
  \And
  Rajarshi Roy $^{1}$
  \And 
  Peng Xu $^{1}$
  \AND
  Chankyu Lee $^{1}$
  \And
  Mohammad Shoeybi $^{1}$ \\
  \\
  $^{1}$ NVIDIA
  \And
  Bryan Catanzaro $^{1}$
}

\begin{document}

\maketitle

\begin{abstract}
In this work, we introduce ChatQA, a suite of models that outperform GPT-4 on retrieval-augmented generation~(RAG) and conversational question answering~(QA).
To enhance generation, we propose a two-stage instruction tuning method that significantly boosts the performance of RAG.
For effective retrieval, we introduce a dense retriever optimized for conversational QA, which yields results comparable to the alternative state-of-the-art query rewriting models, while substantially reducing deployment costs.
We also present the \textsc{ChatRAG Bench}, which encompasses ten datasets covering comprehensive evaluations on RAG, table-related QA, arithmetic calculations, and scenarios involving unanswerable questions.
Our ChatQA-1.0-70B (score: 54.14), built on Llama2, a weaker foundation model than GPT-4, can slightly outperform GPT-4-0613 (score: 53.90) and GPT-4-Turbo-2024-04-09~(score: 54.03) on the \textsc{ChatRAG Bench}, without relying on any synthetic data from OpenAI GPT models.   
Notably, the Llama3-ChatQA-1.5-70B model surpasses the accuracy of GPT-4-Turbo-2024-04-09, achieving a 4.4\% improvement.
To advance research in this field, we open-sourced the model weights, instruction tuning data, \textsc{ChatRAG Bench}, and retriever for the community: \url{https://chatqa-project.github.io/}.
\end{abstract}

\section{Introduction}
Most recently, ChatGPT~\citep{chatgpt} and its follow ups~\citep{gpt4, claude, bard} have led to the paradigm shift of building question answering~(QA) and retrieval-augmented generation~(RAG) system in production and research community.
In particular, the following aspects of the models are preferred:
\emph{i)}~The users can interact with the QA models in a conversational way, thus one can easily raise follow-up questions.
\emph{ii)}~The models are capable of integrating retrieved chunks of evidence in both open-domain or long document settings, where the provided context is much longer than the context window of LLM~\citep[e.g.,][]{claud_100k, xu2023retrieval}.
\emph{iii)}~The generalist models can answer any questions with respect to table, arithmetic calculation in zero-shot manner without dataset-specific fine-tuning, while matching the accuracies of fine-tuned models.
% \zihan{To incorporate these three aspects, we focus on improving model's zero-shot capability on the context-based conversational QA with follow-up questions.}
To this end, we focus on building the state-of-the-art model with all these key capabilities that are essentially important for many real-world applications.

However, building such a model that can match the accuracy of the state-of-the-art proprietary models, e.g., GPT-4~\citep{gpt4}, is still a grand challenge for the research community.
In this work, we introduce \method, a family of open-sourced models that can outperform GPT-4 while utilizing relatively weak open-weights foundation model.
We also open-source our training data, detailing techniques for synthetic data generation, along with alternative human annotation processes aimed at eliminating reliance on OpenAI GPT models for open research purpose.

Specifically, we make the following contributions: 
\vspace{-0.4em}
\begin{enumerate}[leftmargin=3.1em]
    % instruction tuning
    \item We propose a two-stage instruction tuning method and design a dataset curation recipe that can largely enhance LLM's capability of integrating user provided or retrieved context for  conversational QA and RAG tasks. 
    We demonstrate that the proposed instruction tuning method significantly outperforms strong alignment baselines or RLHF-based recipes~(e.g., Llama2-Chat, Llama3-Instruct) on RAG and various conversational QA tasks.
    %
    % retriever
    \item For retrieval, we show that fine-tuning the single-turn QA retriever on human-annotated data or synthetic multi-turn QA dataset works as well as utilizing the state-of-the-art LLM-based query rewriting model, i.e., GPT-3.5-Turbo~\citep{chatgpt}. 
    Our result also highlights the promising direction of utilizing synthetic data generation for training customized retriever.
    %
    % comprehensive evaluation
    \item 
    We introduce \textsc{ChatRAG Bench}, a comprehensive benchmark with ten conversational QA datasets, including five datasets with long documents that need retrieval and three datasets with tabular data and arithmetic calculation.
    We apply ChatQA training recipe on different text foundation models, and show the superb generalization capability of the proposed methods.
    In terms of average score on \textsc{ChatRAG Bench}, our ChatQA-1.0-70B~(54.14) based on Llama2 can outperform  GPT-4-0613~(53.90) and GPT-4-Turbo-2024-04-09~(54.03) without utilizing any synthetic data from ChatGPT models.
    Notably, much smaller Llama3-ChatQA-1.5-8B can perform comparable with GPT-4 models, while Llama3-ChatQA-1.5-70B outperforms GPT-4-Turbo-2024-04-09 by a margin.
    % In human evaluation, \method-70B is slightly worse than GPT-4 in terms of subjective preference.
    %
    \item We study the ``unanswerable'' scenario, where the LLM needs to generate ``cannot answer'' to avoid hallucination.
    We show that incorporating a small amount of ``unanswerable'' samples significantly enhances model's capability to handle it. 
    Our ChatQA-1.0-70B outperforms GPT-3.5-Turbo in this regard, while  has a slight gap compared to GPT-4-0613~(around 3.5\%).
\end{enumerate}

% We organize the rest of the paper as follows.
We discuss related work in \S~\ref{sec:related}.
We introduce the two-stage instruction tuning method and data curation for \method in \S~\ref{sec:chatqa}, and study retrieval in conversational QA in \S~\ref{sec:multi_turn_retrieval}.
We present the experimental setup in \S~\ref{sec:experiment_setup}, results in \S~\ref{sec:results}, and conclude the paper in \S~\ref{sec:conclusion}.

%%%%%%%%%%%%%%%%%%%%%%%%%%%%%%%%%%%%%%%%%%%%%%%%%%%%%%%%%%%%

\section{Related Work}
\label{sec:related}
\subsection{Conversational QA and RAG}
Question answering in a conversational way naturally improves user experiences by addressing follow-up questions. The model can also raise clarification questions for users if necessary, which can reduce hallucination. 
Thus, it becomes the default format of deploying QA models in production~\citep[e.g.][]{chatgpt, bard, claude}.
In contrast to the latest LLM-based generalist solution~\citep[e.g.,][]{chatgpt}, most of the previous studies focus on fine-tuned expert models on specific domains or datasets~\citep{feng2020doc2dial,izacard2021leveraging,chen2022convfinqa,gao2022unigdd,nakamura2022hybridialogue,adlakha2022topiocqa,wu2023inscit}.

In recent years, many conversational QA datasets have been introduced. The models are asked to answer questions based on provided context or documents, which involves retrieval-augmented generation~(RAG) if the provided documents are longer than the context window of LLM.
The provided context or documents can be: \emph{i}) text-only documents from various domains~\citep{feng2020doc2dial,anantha2021open,saeidi2018interpretation,adlakha2022topiocqa,aliannejadi2021building,reddy2019coqa,qu2020open,wu2023inscit,deng2022pacific,guo2021abg,choi2018quac,campos2020doqa}, or \emph{ii}) documents comprising plain text along with tables~\citep{pasupat2015compositional,nakamura2022hybridialogue,chen2022convfinqa}.

\subsection{Retrieval for Multi-Turn QA}
RAG is critically important for conversational QA in open-domain setting, e.g., utilizing update-to-date information from search engine, or when the proprietary documents are longer than the context window of LLM.
The dense retrievers are usually trained to retrieve the top-\emph{k} relevant chunks given a single question~\citep[e.g.,][]{lin2023train, wang2022text, izacard2021contriever}.
In conversational QA, the follow-up questions~(e.g., with pronouns referring to entities mentioned in the previous conversation) may have insufficient information for retrieval, while feeding them along with all of the dialogue history can be redundant, thus leading to sub-optimal results.

\paragraph{Conversational Query Rewriting}
Most of the previous solutions are query rewriting methods. 
The latest turn of question is rewritten to be a standalone query without additional information from previous dialogue history~\citep{vakulenko2021question,ye2023enhancing,mo2023convgqr}, so it can be directly used by retrieval model to retrieve relevant context~\citep{vakulenko2021comparison,mele2021adaptive,raposo2022question,mo2023convgqr}.
%
% Conversational query rewriting methods rewrite the last question turn of a conversation and make it not dependent on the previous dialogue history~\citep{vakulenko2021question,ye2023enhancing,mo2023convgqr}. The rewritten query can be directly used by retrieval model to retrieve relevant context~\citep{vakulenko2021comparison,mele2021adaptive,raposo2022question,mo2023convgqr}. 
Many datasets have been collected to facilitate this line of research~\citep{elgohary2019can,chu2020ask,qu2020open,anantha2021open,brabant2022coqar}, alongside multiple proposed query rewriting methods~\citep{ishii2022can,yu2020few,wu2022conqrr,del2021question,chen2022reinforced,galimzhanova2023rewriting}. 
For example, \citet{wu2022conqrr} and \citet{chen2022reinforced} proposed to use reinforcement learning methods for the query rewriting.
\citet{yu2020few} investigated few-shot generative models like GPT-2 for query rewriting. 
\citet{galimzhanova2023rewriting} studied instruction tuned GPT-3.5-Turbo and showed that it achieved state-of-the-art results for conversational query rewriting.

\paragraph{Fine-tuning Retriever for multi-turn QA}
Some previous work fine-tune a single-turn query retriever on in-domain conversational query and context pairs~\citep{feng2020doc2dial,gao2022unigdd,adlakha2022topiocqa,wu2023inscit}, so it can directly take a concatenation of dialog history and current query as input.
In this work, we focus on the zero-shot evaluation. 
We fine-tune a single-turn query retriever on a high-quality multi-turn dataset. Then, we evaluate zero-shot capability of the fine-tuned retriever on five benchmark datasets.
Surprisingly, we find this simple approach can obtain comparable zero-shot results as the state-of-the-art query rewriting model, i.e., GPT-3.5-Turbo.

\subsection{Instruction Tuning}

% survey \citep{zhang2023instruction}

The goal of instruction tuning is to equip LLMs with the capability to follow natural language instructions~\citep{wei2021finetuned, sanh2021multitask, mishra2021cross,iyer2022opt,du2022glm,ouyang2022training,wang2023far,zhang2023instruction,gao2023retrieval, chung2022scaling, muennighoff2022crosslingual, xu2023wizardlm, wang2022super, zhou2023lima, albalak2024survey}. 
There has been a surge in the development of high-quality instruction tuning datasets, including FLAN~\citep{chung2022scaling}, Self-Instruct~\citep{wang2022self}, unnatural Instructions~\citep{honovich2022unnatural}, Dolly~\citep{DatabricksBlog2023DollyV2}, and OpenAssistant \citep{köpf2023openassistant}.

Although numerous research on instruction tuning has been conducted, a few works focused on improving RAG or context awareness generation for QA.
% context-based and multi-stage instruction tuning. 
\citet{lin2023ra} introduced a retrieval-augmented instruction tuning method, which appends top-\emph{k} retrieved chunks for LLM fine-tuning. 
\citet{wang2023instructretro} applied instruction tuning after retrieval-augmented pretraining.
In contrast, we propose a two-stage instruction tuning method to improve generation with retrieval or provided context. We find that appending top-\emph{k} retrieved chunks for LLM fine-tuning does not help for a wide range of conversation QA tasks~(see \S\ref{subsec:ablation_study_narrative_qa} for details).

% The goal of instruction tuning is to equip LLMs with the capability to comprehend and follow natural language instructions~\citep{sanh2021multitask,mishra2021cross,iyer2022opt,du2022glm,ouyang2022training,wang2023far,zhang2023instruction,gao2023retrieval}. Recently, there has been a surge in the development of high-quality instruction tuning datasets~\citep{sanh2021multitask,chung2022scaling,muennighoff2022crosslingual,xu2023wizardlm,honovich2022unnatural,mishra2022cross,wang2022super,zhou2023lima,köpf2023openassistant}. For example, FLAN datasets~\citep{chung2022scaling}, Self-Instruct~\citep{wang2022self}, unnatural Instructions~\citep{honovich2022unnatural}, Dolly~\citep{DatabricksBlog2023DollyV2}, and OpenAssistant \citep{köpf2023openassistant}.

Similar to the latest work~\citep{zhang2023r}, we demonstrate that adding a small amount of ``unanswerable'' samples in instruction tuning can steer the model to generate ``cannot answer'' output when it is necessary, thus significantly reduce hallucination. 

% After the release of ChatGPT~\citep{chatgpt}, instruction tuning became an indispensable ingredient to build the state-of-the-art dialogue agent which has astonishing zero-shot capability to a wide range of tasks.
%~\citep{almazrouei2023falcon,ding2023enhancing,taori2023alpaca,peng2023instruction,YuLan-Chat,chiang2023vicuna,touvron2023llama2,claude,gpt4,bard,Gemini}.
% The conversational QA capability plays a crucial role in a dialogue agent, yet there has been limited research dedicated to this vital aspect. 

% Despite numerous research work about instruction tuning has been conducted, very few focused on context-based instruction tuning and 
% RA-DIT~\citep{lin2023ra}

\begin{figure*}
    \centering
    \includegraphics[width=0.85\linewidth]{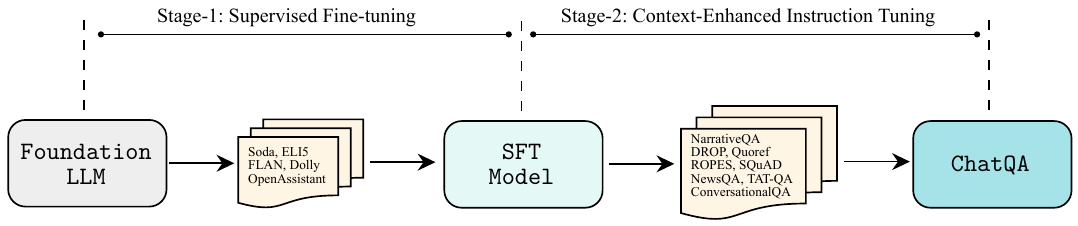}
    % \vspace{-0.5em}
    \caption{Two-stage instruction tuning framework for ChatQA.}
    \label{fig:chatqa}
\end{figure*}

\section{\method}
\label{sec:chatqa}

In this section, we propose a two-stage instruction tuning method for ChatQA. See Figure~\ref{fig:chatqa} for an illustration. 
Our method starts with a pretrained LLM foundation model. 
At stage-1,  we apply supervised fine-tuning (SFT) as in~\citet{ouyang2022training} on a blend of instruction-following and dialog datasets.  
After that, our model exhibits good capability to follow instructions as a conversational agent. However, its capability for contextualized or RAG-based QA remains limited. Hence, we introduce a subsequent stage, called context-enhanced instruction tuning, which is designed specifically for enhancing our model's capability for context-aware or retrieval-augmented generation in conversational QA.

% to incorporate context in the instruction tuning stage.
% In stage-2, 

\subsection{Stage-1: Supervised Fine-tuning}

To construct a large and comprehensive supervised fine-tuning~(SFT) dataset, we follow~\citet{xu2023retrieval}, \citet{wang2023instructretro} and gather a combined set of 128K SFT  samples from high-quality instruction tuning datasets. It consists of 1) a social dialogue dataset Soda~\citep{kim2022soda}, 2) a long-form QA dataset ELI5 containing elaborate answers~\citep{fan2019eli5}, 3) FLAN and chain-of-thought datasets~\citep{wei2022chain,chung2022scaling,longpre2023flan}, 4) LLM synthetic instruction tuning datasets, including Self-Instruct~\citep{wang2022self} and Unnatural Instructions~\citep{honovich2022unnatural}, and 5) a private crowd-sourced conversational dataset, as well as two public human-written conversation datasets: OpenAssistant~\citep{kopf2023openassistant}, and Dolly~\citep{conover2023free}.

We unify the structure of all the SFT data in a conversational format. We first add a ``System'' role at the beginning to set up a general instruction guiding LLM to provide polite and helpful answers. 
We also add ``User'' and ``Assistant'' roles to incorporate instruction and response pairs from the instruction tuning datasets.
We apply fine-tuning using this unified format on an LLM foundation model.

\subsection{Stage-2: Context-Enhanced Instruction Tuning}

To further enhance the model's conversational QA capability over a given context, we conduct a second stage instruction tuning, which integrates contextualized QA datasets into the instruction tuning blend. 
Specifically, the stage-2 instruction tuning datasets consist of a blend of contextualized single-turn QA and conversational QA datasets. 
We present further details for the stage-2 instruction tuning datasets below.

\subsubsection{Human Annotated Data}
%
% To avoid the dependence on OpenAI models (e.g., GPT-3.5-Turbo) to generate the synthetic dataset for building our conversational QA models, 
In addition to the public available datasets, one of the key elements for stage-2 is to obtain a high-quality document-grounded conversational QA dataset.
We create a human-annotated conversational QA (called HumanAnnotatedConvQA) dataset only consisting of 7k dialogues. To build this dataset, we first collected 7k documents covering diverse topics from the Internet. Then, we instruct the annotator to act both as a curious user asking questions (and follow-up questions) about the document and as an agent to give the answers. We create a multi-turn conversation for each document, resulting in a total of 7k conversational QA dialogues with an average of 5 user-agent turns per dialogue. {Details of data collection guideline can be found in Appendix~\ref{appendix:data_collection}.}

To reduce hallucinated answers in unanswerable cases, we aim to empower our model to explicitly indicate it when the answer cannot be found within the given context. 
To obtain these unanswerable data samples, we requested annotators to identify all parts of the context locations that are relevant to the user's question. Hence, it enabled us to construct unanswerable scenarios by deleting the text from the corresponding locations in the context. 
After deleting the relevant text to the question, we use a sentence, ``\texttt{Sorry.~I cannot find the answer based on the context}'', as the response for the unanswerable questions. 
Finally, we construct another 1.5k user-agent turns with unanswerable annotations, which provides a good trade-off of answerable and unanswerable cases~(see \S\ref{sec:unanswerable} for details).
% (We find that 2k unanswerable samples will be more or less an optimal number). 

\subsubsection{Synthetic Data Generation}
To validate the quality of the HumanAnnotatedConvQA, we leverage GPT-3.5-Turbo to generate a synthetic conversational QA dataset given its powerful instruction-following and text generation capability.
Note that large-scale synthetic data for conversational QA has also been explored in \citet{dai2022dialoginpainting}. 
In this work, we focus on midsize high-quality synthetic data for LLM fine-tuning.

The instruction for GPT-3.5-Turbo comprises three parts: {1})~system role guiding the model to provide helpful answers, {2})~examples of conversational QA indicating the required data types, and {3})~a document that directs the model to generate conversational QA based on its content.
We collect 7k documents (average $\sim$1k words per document) from common crawl, which cover a wide range of domains. Each document is used for generation of a single conversational QA sample, which leads to a total of 7k multi-turn QA dialogues with an average of 4.4 user-agent turns per dialogue (called SyntheticConvQA).

Similar to the HumanAnnotatedConvQA, we construct another 1.5k user-agent turns with unanswerable annotations in this synthetic dataset. 
Since there are no annotations of the context location for the agent's answer, we construct synthetic unanswerable samples from SyntheticConvQA.
Specifically, we first cut the document (for each dialogue) into different chunks. Then, we consider it as a valid unanswerable sample only when there are chunks that have ``high overlaps'' with the agent's answer to be removed and the rest of the chunks show ``low overlaps'' with the agent's answer. We use the 4-gram recall score between each chunk and the agent's answer (which measures the ratios of the answer's 4-gram phrases are within each chunk) as the metric to measure their overlaps, and consider it higher than 0.5 as ``high overlaps'' and lower than 0.1 as ``low overlaps''.

% consider it as a valid unanswerable example only if the removed context has a high overlap with the original answer, and the rest of the context shows low overlap with the original answer to ensure the data quality of unanswerable cases.
 
% to ensure the data quality, we consider it as a valid unanswerable example only if the removed context has a high overlap with the original answer, and the rest of the context shows low overlap with the original answer.

\subsubsection{Training Blends}
In this part, we introduce the details of training blends for stage-2 instruction tuning.
% In addition to the above mentioned synthetic and human-annotated datasets, we incorporate a proprietary conversational QA dataset (called ProprietaryConvQA) consisting of 1.2K dialogs to further enhance our model's capability in conversational QA. 
To boost the QA capability in handling tabular documents and arithmetic calculation, we add the TAT-QA dataset~\citep{zhu2021tat} which contains both elements. In addition, we integrate contextualized single-turn QA datasets to further strengthen the QA capability of our model. We also retain the stage-1 SFT dataset in the training blend to maintain the model's instruction-following capability.

Finally, the training blend for stage-2 consists of:
1)~A conversational QA dataset: HumanAnnotatedConvQA or SyntheticConvQA,~\footnote{Unless specified otherwise, the experiments use HumanAnnotatedConvQA as the default setting.}
2)~single-turn QA datasets: DROP~\citep{dua2019drop}, NarrativeQA~\citep{kovcisky2018narrativeqa}, Quoref~\citep{dasigi2019quoref}, ROPES~\citep{lin2019reasoning}, SQuAD1.1~\citep{rajpurkar2016squad}, SQuAD2.0~\citep{rajpurkar2018know}, NewsQA~\citep{trischler2017newsqa}, TAT-QA~\citep{zhu2021tat}, and 3)~all of SFT datasets from stage-1. 
As for the training of Llama3-ChatQA-1.5, we additionally incorporate HybriDial~\citep{nakamura2022hybridialogue} and our collected around 2K QA pairs within the financial domain to further improve our model's capability in tabular understanding and arithmetic calculations. 
% and ProprietaryConvQA, 
To ensure there is no data contamination, we have verified that the documents in test benchmarks do not exist in our training data.

We follow the similar template as in stage-1 to unify all single-turn QA and conversational QA datasets. The differences are in two parts: 1) Following the system role, we append relevant context for the single-turn question or the multi-turn conversation, and 2) Just before the single-turn question or multi-turn conversation, we integrate further instruction  based on the answer types of different QA datasets (e.g., short answer, long answer, arithmetic calculation). We use the format for SFT dataset from stage-1.~\footnote{Details of the templates for both stage-1 and stage-2 instruction tuning as well as the synthetic data generation can be found in the Appendix~\ref{appendix:chatqa_instruction_tuning}.}

% \subsubsection{Prompt Templates}

% \subsubsection{Implementation Details}
% We use quite cockatoos datasets for the SFT training.
% 1. Data blending of quite cockatoos datasets for SFT
% 2. Data blending of QA datasets for ChatQA
% 3. Prompt format

\begin{figure}
    \centering
    \includegraphics[width=0.6\linewidth]{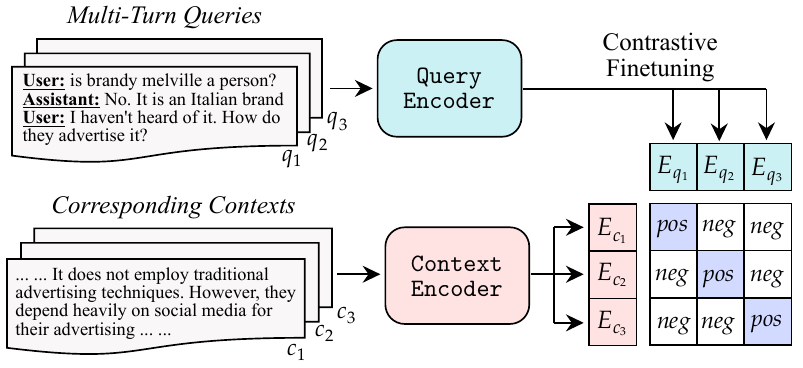}
    % \vspace{-0.6em}
    \caption{Illustration of fine-tuning retriever for multi-turn QA.}
    \label{fig:retrieval}
\end{figure}

\section{Retrieval for Multi-Turn QA}
\label{sec:multi_turn_retrieval}
In conversational QA tasks, when a document becomes too lengthy to feed directly into LLMs, a retriever that can handle conversational queries becomes essential. 
This conversational retriever encodes the concatenation of the dialogue history and the current query, and then retrieve relevant context from documents. After that, only the relevant context will be used as inputs for LLMs. 
The state-of-the-art retrievers, e.g., Dragon~\citep{lin2023train}, are optimized for single-turn queries, resulting in a limited generalization capability for multi-turn conversational queries.
In Figure~\ref{fig:retrieval}, we depict our retriever fine-tuning method to alleviate this issue. We propose to use conversational query and context pairs for further fine-tuning a single-turn retriever to better cope with conversational inputs. 
% In Figure~\ref{fig:retrieval}, we illustrate the differences between single-turn retriever and multi-turn retriever.

An alternative solution is conversational query rewriting method which uses a query rewriter to rewrite the current question based on the conversational history. 
The rewritten query is then directly used as the input to a single-turn query retriever for retrieving relevant context.
In addition to the embedding and search cost, the query rewriting model introduces a large amount of extra computational expense to generate the rewritten query.

% We propose to use 
% There are two straightforward methods for building a multi-turn retriever based on a single-turn one. First, we can use conversational query and context pairs for further fine-tuning the single-turn retriever to better cope with conversational inputs (multi-turn fine-tuning). Second, given the conversational history, we can use a query rewriter to rewrite the current question, which can be directly utilized by a single-turn retriever (conversational query rewriting). 
% In this section, we compare these two methods in a zero-shot setting across five datasets.
% introduce how we use these two methods to build multi-turn retrievers based on a state-of-the-art single-turn retriever Dragon, and we compare the these two methods in a zero-shot setting across five datasets.

% \subsection{Comparisons}

\subsection{Fine-tuning Retriever for Multi-turn QA}
To build a high-quality fine-tuning dataset, we leverage the conversational QA dataset from either the HumanAnnotatedConvQA or the SyntheticConvQA to construct conversational query and context pairs.

For the HumanAnnotatedConvQA, we directly take the annotations of the conversational query and context pairs, and use them to further fine-tune a single-turn query retriever.
For the SyntheticConvQA, we first cut each document in the conversational QA dataset into different chunks. Then, we calculate the 4-gram recall score between agent's answer and each chunk. After that, we consider the chunk that has the highest recall score as the gold chunk for the current user's question. Finally, the constructed conversational query and context pairs are used to fine-tune a single-turn query retriever.

\subsection{Conversational Query Rewriting}
To build powerful conversational query rewriting model, we take GPT-3.5-Turbo as the rewriter given that \citet{galimzhanova2023rewriting} demonstrated the state-of-the-art query rewriting results using GPT-3.5-Turbo. 
Similar to~\citet{galimzhanova2023rewriting}, we not only provide GPT-3.5-Turbo with the rewriting task instruction, but also give it few-shot rewriting examples to enhance the quality of rewriting results.
More details can be found in Appendix~\ref{appendix:query_rewriting_prompts_GPT-3.5-Turbo}.
% The prompt we provide to GPT-3.5-Turbo includes the instructions of the CQR task, few-shot CQR examples, and current conversational.\footnote{More details of the GPT-3.5-Turbo as question rewriter is in the Appendix~\ref{appendix:baselines}.}

\subsection{Comparisons}
\label{subsec:retrieval_comparison}
In Table~\ref{tab:retrieval_main}, we compare the query rewriting and fine-tuning methods across five datasets in the zero-shot setting. More details about these datasets can be found in~\S\ref{subsec:long_benchmarks}. We conduct experiments on a state-of-the-art retriever, Dragon~\citep{lin2023train}, and a strong unsupervised retriever, E5-unsupervised~\citep{wang2022text}, which is  not finetuned on MS MARCO~\citep{nguyen2016ms}. 
% We consider it is a fair comparison between these two methods because they both leverage the powerful text generation and instruction following capability of GPT-3.5-Turbo.
In terms of the experiments on Dragon, we find that fine-tuning performs marginally worse  than query rewriting in average top-1 recall by 1.74\%, while it achieves better results on average top-5 recall by 0.54\%.
It demonstrates the effectiveness of the fine-tuning approach for the conversational retrieval.
In addition, we observe that the results are comparable between using HumanAnnotatedConvQA and SyntheticConvQA for fine-tuning. This highlights that our human-annotated dataset is in high-quality, and we \emph{do not rely on} ChatGPT models for building the state-of-the-art multi-turn query retriever.

Surprisingly, fine-tuning performs significantly better than rewriting on E5-unsupervised. We conjecture that since E5-unsupervised does not use human-annotated query and context pairs in the pre-training stage, it leads to weak generalization for the high-quality rewritten query. In contrast, using a high-quality dataset to fine-tune E5-unsupervised brings a giant boost, with more than a 15\% improvement on both average top-1 and top-5 recall scores.

Therefore, fine-tuning a good single-turn retriever on high-quality conversational query context pairs performs on par with leveraging the state-of-the-art rewriter. However, rewriting method requires extra computational time for autoregressive generation process and probably also API cost for using powerful models like GPT-3.5-Turbo. In contrast, our proposed multi-turn fine-tuning bypasses these issues. 
For the QA evaluations across these five datasets, we consistently use the retrieved top-5 results from the fine-tuning approach for all the QA models. We put more results on comparisons between rewriting and fine-tuning methods in the Appendix~\ref{appendix:More_results_for_Multi-Turn QA Retrieval}.

\begin{table*}[]
\renewcommand{\arraystretch}{1.03}
\centering
\begin{adjustbox}{width={1.03\textwidth},totalheight={\textheight},keepaspectratio}
\begin{tabular}{lcccccccccccc}
\toprule
\multirow{2}{*}{Models}                                                                                                         & \multicolumn{2}{c}{Average} & \multicolumn{2}{c}{Doc2Dial} & \multicolumn{2}{c}{QuAC} & \multicolumn{2}{c}{QReCC} & \multicolumn{2}{c}{TopiOCQA} & \multicolumn{2}{c}{INSCIT} \\ \cmidrule(lr){2-3} \cmidrule(lr){4-5} \cmidrule(lr){6-7} \cmidrule(lr){8-9} \cmidrule(lr){10-11} \cmidrule(lr){12-13}
& top-1         & top-5        & top-1       & top-5      & top-1       & top-5   & top-1        & top-5     & top-5*       & top-20*       & top-5*        & top-20*        \\ \midrule
\citet{adlakha2022topiocqa} & - & - & - & - & - & - & - & - & - & 70.40$^\triangle$ & - & - \\
\citet{wu2023inscit} & - & - & - & - & - & - & - & - & - & - & - & 71.10$^\triangle$ \\ \midrule
E5-unsupervised$^\dagger$~\citep{wang2022text} & 31.56 & 59.22 & 23.02 & 55.33 & 43.49 & 77.68 & 44.71 & 84.99 & 26.25 & 37.67 & 20.32 & 40.44  \\
E5-unsupervised + Rewrite$^\ddagger$ & 33.23 & 61.02 & 25.56 & 58.00 & 46.00 & 80.01 & 45.50 & 85.89 & 27.58 & 39.15 & 21.53 & 42.04 \\
E5-unsupervised + Fine-tune$^\dagger$ & \textbf{47.79} & \textbf{75.00} & 45.28 & 80.96 & 46.52 & 80.74 & 53.37 & 89.91 & 41.01 & 51.07 & 52.79 & 72.31 \\ \midrule
Dragon$^\dagger$~\citep{lin2023train}   & 46.29       & 73.09    & 43.33         & 75.61        & 56.80        & 82.86      & 46.17       & 81.96       & 57.68        & 78.80   & 27.49 & 46.22           \\
Dragon + Rewrite$^\ddagger$     & \textbf{54.46}       & 80.13   & 47.60          & 80.60         & 47.10        & 77.15      & 51.73       & 85.78       & 73.07        & 88.19  & 52.79 & 68.92           \\
Dragon + Fine-tune$^\dagger$  &  52.72 & \textbf{80.67} & 48.94 & 83.01 & 52.64 & 81.95 & 50.73 & 87.17 & 67.86 & 86.28 & 43.43 & 64.94        \\ 
- SyntheticConvQA$^\diamondsuit$ &  52.98       & \textbf{81.15}    & 48.64         & 83.47        & 54.75       & 83.23      & 49.63       & 86.70        & 64.48        & 85.24 & 47.41 & 67.13 \\
\bottomrule
\end{tabular}
\end{adjustbox}
\caption{Retrieval results across five multi-turn QA datasets with the average top-1 and top-5 recall scores. 
Compared to rewriting, fine-tuning performs much better on E5-unsupervised and is comparable on Dragon.
*Since the average context length in TopiOCQA and INSCIT is smaller than in other datasets, we report top-5 and top-20 to roughly match the context lengths of top-1 and top-5, respectively, in those datasets. $^\dagger$The inputs for these two models are a concatenation of the dialogue history and the current query. $^\ddagger$The input for this model is the rewritten query. 
$^\diamondsuit$denotes that the HumanAnnotatedConvQA dataset is replaced with the SyntheticConvQA for fine-tuning. $^\triangle$The numbers are not apple-to-apple comparison (e.g., they use the training set for fine-tuning).}
\label{tab:retrieval_main}
\end{table*}

\section{Experimental Setup}
\label{sec:experiment_setup}
In this section, we present the details of our experimental setup for the conversational question answering task.

\subsection{Baselines}
We develop ChatQA models based on our in-house GPT-\{8B, 22B\} base models (pretrained with 3.5 trillion tokens), Llama2-\{7B, 13B, 70B\} base models~\citep{touvron2023llama2}, and Llama3-\{8B, 70B\} base models~\citep{llama3}. We compare ChatQA models against \textbf{Llama2-Chat-\{7B, 13B, 70B\}} and \textbf{Llama3-Instruct-\{8B, 70B\}}, which are shown to possess strong instruction following and conversational QA capabilities~\citep{touvron2023llama2}.
In addition, we compare against a powerful RAG model, \textbf{Command R+}~\citep{command_r_plus2024}, which has 104 billion parameters, as well as three very strong OpenAI models: \textbf{GPT-3.5-Turbo-0613}, \textbf{GPT-4-0613}, and \textbf{GPT-4-Turbo-2024-04-09}.
For fair comparison, when retrieval is needed, we use the same top-\emph{k} retrieved chunks from our best retriever as the context for all baselines and our ChatQA models.
Note that we have carefully tuned the instructions for all the baselines to ensure they achieve as good as possible results.\footnote{The prompts for these baselines can be found in Appendix~\ref{appendix:benchmarks}.}

% In addition, we adjust the prompts for GPT-4

% \subsection{Evaluation Setup}
% \subsubsection{Retrieval Benchmarks}
% We need datasets that have long documents for each conversation to evaluate the quality of retrievers. 
% The average document size per conversation for most conversational QA datasets is less than 1K words, which can be directly used as inputs in LLMs without the need of retrieval. Originally, we only found Doc2Dial~\citep{feng2020doc2dial} and TopiOCQA~\citep{adlakha2022topiocqa} have long documents for each conversation. 
% To make the retrieval evaluation more comprehensive, we extended the documents of QuAC~\citep{choi2018quac} and QReCC~\citep{anantha2021open} by crawling the text from the corresponding URLs for each conversation. Eventually, we evaluate retrieval on four datasets: Doc2Dial, TopiOCQA, QuAC and QReCC.
%% TODO
%% Add more details about how we extend the documents of QuAC and QReCC (Appendix)
%% Statistics of these datasets

\subsection{\textsc{ChatRAG Bench}: Evaluation Benchmarks}
To evaluate the model's capability on conversational QA and RAG, we construct \textsc{ChatRAG Bench}, a collection of 10 datasets covering a wide range of documents and question types, which require models to generate responses from (retrieved) context, comprehend and reason over tables, conduct arithmetic calculations, and indicate when questions cannot be found within the context.

\subsubsection{Long Document Datasets}
\label{subsec:long_benchmarks}

We collect five conversational QA datasets with long documents which cannot be directly fitted into LLMs with a sequence length of 4K or 8K tokens. Hence, we run our best multi-turn query retriever to get the top-n relevant chunks as the inputs (experiments can be found in \S\ref{subsec:retrieval_comparison}).
\begin{itemize}
    \item Doc2Dial (D2D)~\citep{feng2020doc2dial} is a document-grounded conversational QA dataset covering four domains: DMV, SSA, VA, and Student Aid. Each sample comprises a dialogue where a user poses queries regarding the document, and an agent responds those questions. The average document length is around 101K words. 
    \item QuAC~\citep{choi2018quac} is based on Wikipedia documents. Originally, the document is short. Since each dialogue is linked to multiple Wikipedia URLs, we extract text from these links to increase the document size to approximately an average of 15K words. It contains unanswerable cases where answers cannot be found within the given context.
    \item QReCC~\citep{anantha2021open} is an open-domain conversational QA dataset across multiple sources. 
    Similar to QuAC, each dialogue also has corresponding URLs. We extract text from those URLs to construct the documents. In the end, the average document size is around 5K words, with a maximum document size of 20K words.
    \item TopiOCQA (TCQA)~\citep{adlakha2022topiocqa} is grounded on the whole Wikipedia. It incorporates topic switching and requires the agent to search the entire Wikipedia for answers to user questions. 
    % For the Wikipedia corpus chunking, we follow the original segmentation provided by~\citet{adlakha2022topiocqa}.
    \item INSCIT~\citep{wu2023inscit} is also grounded on the whole Wikipedia. It studies the case where user questions are under-specified and require clarification.
    
\end{itemize}

For Doc2Dial, QuAC, and QReCC, we segment documents into around 300-word chunks, and we retrieve top-5 relevant chunks as context for each user question. For TopiOCQA and INSCIT, we follow their original segmentation, resulting in smaller chunks. Hence, we retrieved top-20 chunks to obtain similar context length to the first three datasets.

\subsubsection{Short Document Datasets}
To increase the diversity of document lengths, we collect five conversational QA datasets with short documents (less than 1.5K words). On average, 1 word will be tokenized into ~1.5 tokens. Hence, the document can be directly fitted into LLMs with a sequence length of 4K tokens.

\begin{itemize}
    \item CoQA~\citep{reddy2019coqa} is a conversational QA dataset with each dialogue grounded on a short passage. The answers are generally short, and the passages cover a wide range of domains like children's stories, literature, mid/high school exams, news, Wikipedia.
    \item DoQA~\citep{campos2020doqa} covers three domains: cooking, travel, and movies collected from active Stack Exchange\footnote{\url{https://stackexchange.com/}} forums. The dataset contains unanswerable cases where answers cannot be found within the given document.
    \item ConvFinQA (CFQA)~\citep{chen2022convfinqa} is based on the Financial domain. Each document contains a single financial report table along with relevant text surrounding the table. This dataset involves arithmetic calculation and complex numerical reasoning. 
    \item SQA~\citep{pasupat2015compositional} is grounded on documents which contain only a single Table without any surrounding text. The documents are collected from Wikipedia, and the questions are highly compositional, which requires the model with robust table comprehension ability to give correct answers.
    \item HybriDial (HDial)~\citep{nakamura2022hybridialogue} is a conversational QA dataset grounded on documents containing both Wikipedia tabular and textual data. The questions are complex which requires reasoning over the documents.
\end{itemize}
% \vspace{-1.5cm}
Over all the 10 datasets, ConvFinQA, SQA, and HybriDial datasets contain tabular data in the documents, while documents of the remaining datasets are text-only.\footnote{Details of these benchmark datasets are in the Appendix~\ref{appendix:benchmarks}.}

\subsection{Evaluation Metrics}
Given that F1 score is the most commonly used automatic metric to assess QA models, we use it for all datasets except for ConvFinQA. In ConvFinQA, we follow~\citet{chen2022convfinqa} to use exact match metric since the answers in ConvFinQA are about extracting numbers from documents as well as arithmetic calculations. Hence, the answer only makes sense when it is exactly the same as the answer. When models generate the arithmetic formula, we will calculate its final result based on a calculator and compare it with the gold answer.

In addition, we also conduct human evaluations to assess the correctness of generated answers between our model and GPT-4-0613.

% Aside from long document datasets: Doc2Dial, TopiOCQA, QuAC, QReCC, and INSCIT, we further evaluate our model on another five datasets having short documents: DoQA~\citep{campos2020doqa}, CoQA~\citep{reddy2019coqa}, SQA~\citep{pasupat2015compositional}, ConvFinQA~\citep{chen2022convfinqa}, and HybriDialogue~\citep{nakamura2022hybridialogue}. For short document datasets, we directly use the whole document as the context inputs to the LLMs. Among these ten datasets, six of them are conversations over text only documents: Doc2dial, QuAC, QReCC, TopiOCQA, CoQA, and DoQA. The rest of three datasets (SQA, ConvFinQA, and HybriDialogue) have tabular data in the conversation documents, which allows us to evaluate the Tabular comprehension ability of our model\footnote{More details of these benchmarks can be found in Appendix~\ref{appendix:benchmarks}.}.

% Might need to add more content to justify this evaluation.

\begin{table*}[]
\renewcommand{\arraystretch}{1.03}
\centering
\begin{adjustbox}{width={1.02\textwidth},totalheight={\textheight},keepaspectratio}
\begin{tabular}{lcccccccccccc}
\toprule
Models         & \begin{tabular}[c]{@{}c@{}}Avg. w/o\\HDial \end{tabular}    & \begin{tabular}[c]{@{}c@{}}Avg.\\All\end{tabular}  & D2D & QuAC  & QReCC & CoQA  & DoQA  & CFQA & SQA   & TCQA & HDial & INSCIT \\ \midrule
Llama2-Chat-7B & 37.96  & 38.86 & 33.27 & 25.83 & 46.02 & 72.28 & 33.15 & 36.58 & 26.14 & 36.68 & 47.02 & 31.67  \\
Llama2-Chat-70B & 44.64   & 45.21   & 36.87    & 32.47 & 49.40  & 80.41 & 38.97 & 46.85     & 37.62 & 44.31    & 50.35      & 34.88  \\ 
Llama3-Instruct-8B & 46.76 & 46.70 & 33.59 & 33.60 & 45.70 & 75.26 & 37.33 & 58.05 & 59.72 & 44.96 & 46.20 & 32.59 \\
Llama3-Instruct-70B & 52.95 & 52.52 & 37.88 & 36.96 & 51.34 & 76.98 & 41.24 & 76.60 & 69.61 & 49.72 & 48.59 & 36.23  \\
\midrule
\method-1.0-7B & 46.96 & 47.71 & 37.88 & 29.69 & 46.97 & 76.61 & 41.57 & 51.61 & 61.87 & 45.45 & 54.51 & 30.96 \\ 
\method-1.0-70B & 53.89  &  54.14 & 38.90 & 41.82 & 48.05 & 78.57 & 51.94 & 73.69 & 69.14 & 50.98 & 56.44 & 31.90  \\ 
Llama3-\method-1.5-8B & 53.99 & 55.17 & 39.33 & 39.73 & 49.03 & 76.46 & 49.60 & 78.46 & 73.28 & 49.96 & 65.76 & 30.10 \\ 
Llama3-\method-1.5-70B & \textbf{57.14} & 58.25 & 41.26 & 38.82 & 51.40 & 78.44 & 50.76 & 81.88 & 83.82 & 55.63 & 68.27 & 32.31 \\ \midrule
Command R+ & 51.40 & 50.93 & 33.51 & 34.16 & 49.77 & 69.71 & 40.67 & 71.21 & 74.07 & 53.77 & 46.70 & 35.76 \\
GPT-3.5-Turbo-0613 & 50.69 & 50.37   & 34.83    & 37.17 & 50.46 & 79.33 & 41.11 & 73.15     & 60.63 & 44.30     & 47.42      & 35.27  \\
GPT-4-0613 &   54.35   & 53.90   & 34.16    & 40.29 & 52.01 & 77.42 & 43.39 & 81.28     & 79.21 & 45.09    & 49.81      & 36.34 \\
GPT-4-Turbo-2024-04-09 & \textbf{54.72} & 54.03 & 35.35 & 40.10 & 51.46 & 77.73 & 41.60 & 84.16 & 79.98 & 48.32 & 47.86 & 33.75 \\
\bottomrule
\end{tabular}
\end{adjustbox}
% \vspace{-0.15cm}
\caption{Zero-shot results on \textsc{ChatRAG Bench}. ChatQA-1.0 are built on Llama2 base models, while Llama3-ChatQA-1.5 are built on Llama-3 base models. Note that Llama3-ChatQA-1.5 used training samples from the HDial dataset. Hence, we also compare average scores excluding HDial for the fair comparison.}
\label{tab:cqa_main}
\end{table*}

\begin{table*}[]
\renewcommand{\arraystretch}{1.03}
\centering
\begin{adjustbox}{width={1.0\textwidth},totalheight={\textheight},keepaspectratio}
\begin{tabular}{lccccccccccc}
\toprule
Models             & Avg. & D2D & QuAC  & QReCC & CoQA  & DoQA  & CFQA & SQA   & TCQA & HDial & INSCIT \\ \midrule
\method-1.0-70B    &  \textbf{54.14} & 38.90 & 41.82 & 48.05 & 78.57 & 51.94 & 73.69 & 69.14 & 50.98 & 56.44 & 31.90  \\
- SyntheticConvQA$^\diamondsuit$ & \textbf{54.08}   & 39.19    & 38.33 & 48.73 & 79.83 & 48.65 & 76.44     & 68.63 & 51.30     & 55.68      & 33.98  \\
- w/o stage-1$^\dagger$  &  52.18 & 38.43 & 37.52 & 46.08 & 73.51 & 49.42 & 72.15 & 72.08 & 51.28 & 50.74 & 30.56  \\
- w/o stage-2$^\dagger$ & 43.22   & 34.42    & 25.65 & 41.88 & 73.04 & 28.21 & 46.64     & 58.90  & 37.20     & 55.52      & 30.71 \\
- w/o single-turn*  & 52.25   & 38.30    & 37.89 & 47.08 & 76.74 & 46.43 & 72.42     & 67.41 & 49.85     & 53.16      & 33.18  \\  
- w/o ConvQAData$^\ddagger$ & 48.97 & 35.47 & 28.19 & 41.68 & 75.80 & 31.68 & 73.83 & 68.05 & 45.90 & 58.40 & 30.70 \\ \bottomrule
\end{tabular}
\end{adjustbox}
% \vspace{-0.15cm}
\caption{Ablation studies on ChatQA training. $^\diamondsuit$ denotes that the HumanAnnotatedConvQA is replaced with the SyntheticConvQA. $^\dagger$ denotes that the stage-1 (SFT) or stage-2 tuning is removed from the ChatQA training. * denotes that the single-turn QA datasets are removed from the stage-2 training blends. $^\ddagger$ denotes that the HumanAnnotatedConvQA is removed from the stage-2 training blends. In both ``w/o stage-1'' and ``w/o single-turn'' settings, the SyntheticConvQA data is used.}
\label{tab:cqa_main_ablation_study}
\end{table*}

\section{Results}
\label{sec:results}

\subsection{Main Results}

\paragraph{Overview}
In Table~\ref{tab:cqa_main}, we compare different model variants and OpenAI models across 10 conversational QA datasets. ChatQA-1.0-7B/70B greatly surpass Llama2-Chat counterparts, and Llama3-ChatQA-1.5-8B/70B greatly outperform Llama3-Instruct counterparts. For example, ChatQA-1.0-70B outperforms Llama2-Chat-70B by an average score of 8.93, and Llama3-ChatQA-1.5-70B surpasses Llama3-Instruct-70B by an average score of 4.19 (Avg. w/o HDial).
In addition, ChatQA-1.0-70B slightly outperforms GPT-4-0613 and GPT-4-Turbo, and greatly surpasses Command R+, a powerful RAG model that has 104 billion parameters, as well as GPT-3.5-Turbo (Avg. All). 
Compared to state-of-the-art OpenAI models (i.e., GPT-4-0613 and GPT-4-Turbo), Llama3-ChatQA-1.5-8B achieves comparable results, and Llama3-ChatQA-1.5-70B greatly outperforms both of them.
\footnote{We place full results (e.g., models based on Llama2-13B, and our in-house 8B/22B models) in Appendix~\ref{appendix:chatrag_bench_all}}.

\paragraph{Importance of Stage-1 SFT and Stage-2 Tuning}
In Table~\ref{tab:cqa_main_ablation_study}, we find that removing the stage-1 SFT from the ChatQA training makes the average score drops 1.9 (from 54.08 to 52.18). Aside from the SQA, removing stage-1 makes the model consistently perform worse on other datasets. Results indicate that stage-1 still plays an important role. It enhances the instruction-following capability of ChatQA which is beneficial for stage-2 tuning. In addition, we find that removing ChatQA stage-2 tuning makes the average scores greatly drop by 10.92 (from 54.14 to 43.22). This is because the stage-2 tuning (i.e., context-enhanced instruction fine-tuning) enables the model to learn how to effectively leverage information from retrieved or relevant context.
%, while it is not as crucial as stage-2, given that removing stage-2 (Llama2-70B-SFT) makes the model perform significantly worse.

% \paragraph{Importance of Stage-2 Tuning}
% In Table~\ref{tab:cqa_main_ablation_study}, we find that removing ChatQA stage-2 tuning makes the average scores greatly drop (from 54.14 to 43.22). 
% In terms of averaged scores, Llama2-Chat models surpass SFT model counterparts by a small margin, while our ChatQA models achieve around or over 10 point of absolute improvement over SFT and Chat counterparts. For example, ChatQA-1.0-13B improves on Llama2-13B-SFT and Llama2-13B-Chat by 13.17 (from 37.69 to 50.86) and 10.52 (from 40.34 to 50.86), respectively. This is because context-enhanced instruction fine-tuning enables the model to learn how to effectively extract useful information from retrieved or relevant context.

\paragraph{Effectiveness of Single-Turn Data}
% Effectiveness of single-turn QA datasets in the training blend
To investigate how single-turn QA datasets affect model's multi-turn QA capability, we conduct an ablation study by removing them from the ChatQA-1.0-70B training blends in stage-2. 
As shown in Table~\ref{tab:cqa_main_ablation_study}, incorporating single-turn QA datasets in the stage-2 training blends generally make the scores increase across all benchmark datasets, leading to an average improvement of 1.83 score. Interestingly, we observe improvement in ConvFinQA, SQA, and HybriDial (table-based datasets), despite the added single-turn QA datasets not having tabular data in the documents.
These results align with our intuitions. Adding single-turn QA datasets improves the model's capability to effectively leverage relevant context for the answers, resulting in better scores on \textsc{ChatRAG Bench}.

\paragraph{Effectiveness of Conversational QA Data}
We further explore the how conversational QA data affect the model's multi-turn QA capability by removing HumanAnnotatedConvQA data from the ChatQA stage-2 training blends.
As illustrated in Table~\ref{tab:cqa_main_ablation_study}, ``w/o ConvQAData'' makes the results significantly worse than ChatQA-1.0-70B (average scores degrading from 54.08 to 48.97). We observe large degradation in datasets with text-based documents, such as QuAC, QReCC, and DoQA. However, the degradation in datasets with table-based documents (e.g., ConvFinQA, SQA) are small, thanks to having TAT-QA~\citep{zhu2021tat} in the training blends.

% Interestingly, ``w/o ConvQAData'' can still outperform Llama2-70B-SFT (48.97 vs. 43.22) and Llama2-70B-Chat (48.97 vs. 45.21). This further demonstrates the effectiveness of our proposed stage-2 tuning, which enhances the model's capability to handle the contextualized QA task.

% \vspace{-1em}

\paragraph{Human Annotated Data vs. GPT-3.5-Turbo Synthetic Data}
% To avoid the dependence on OpenAI models (e.g., GPT-3.5-Turbo) to generate synthetic datasets for building our conversational QA models, we create a human-annotated conversational QA dataset consisting of 5k dialogues. To build this dataset, we first collected 5k documents covering diverse topics from Internet. Then, we instruct the annotator to act both as a curious user asking questions (and follow-up questions) about the document and as a agent to give the answers. We create a multi-turn conversation for each document, resulting in 5k conversational QA dialogues in total. 
% Similar to GPT-3.5-Turbo synthetic data, we also synthetically create unanswerable cases from the annotations. We requested annotators to provide context location from which the answer was produced. Hence, it enabled us to construct unanswerable cases for our ChatQA fine-tuning by deleting the text from the corresponding location in the context.

In Table~\ref{tab:cqa_main_ablation_study}, we also compare our ChatQA models using the 7k GPT-3.5-Turbo synthetic dataset (SyntheticConvQA) and our collected 7k human-annotated dataset (HumanAnnotatedConvQA). First, we find that both achieve comparable results in terms of average scores, which suggests that we do not need to rely on synthetic data from OpenAI models to build the state-of-the-art conversational QA models. Second, we find that using human-annotated data achieved significant improvements on QuAC and DoQA datasets. This can be attributed to the fact that the human-annotated data have higher quality on unanswerable cases which exists in QuAC and DoQA datasets. Eventually, it leads to the overall improvements on these two datasets. Detail results and analyses on unanswerable cases can be found in \S\ref{sec:unanswerable}. 

\begin{table}[t]
\renewcommand{\arraystretch}{1.03}
\centering
\begin{adjustbox}{width={0.45\textwidth},totalheight={\textheight},keepaspectratio}
\begin{tabular}{lccc}
\toprule
           & Ours Win & Tie     & GPT-4 Win \\ \midrule
Average    & 13.81\%  & 69.09\% & 17.10\%  \\ \midrule
Doc2Dial   & 14.29\%  & 68.00\% & 17.71\%  \\
QuAC       & 11.67\%  & 73.33\% & 15.00\%  \\
QReCC      & 11.11\%  & 77.22\% & 11.67\%  \\
CoQA       & 7.78\%   & 80.00\% & 12.22\%  \\
DoQA       & 22.78\%  & 57.78\% & 19.44\%  \\
ConvFinQA  & 16.67\%  & 67.78\% & 15.55\%  \\
SQA        & 11.11\%  & 61.67\% & 27.22\%  \\
TopiOCQA   & 19.31\%  & 60.69\% & 20.00\%  \\
HybriDial & 7.78\%   & 78.33\% & 13.89\%  \\
INSCIT     & 15.56\%  & 66.11\% & 18.33\%  \\ \bottomrule
\end{tabular}
\end{adjustbox}
\vspace{.1cm}
\caption{Human evaluation (A/B testing) comparing our ChatQA-1.0-70B to GPT-4-0613 on \textsc{ChatRAG Bench}. In terms of average scores, our model and GPT-4 are tie most of the time (69.09\%), and GPT-4-0613 achieves slightly higher win rate ($\sim$3.3\%) than ours.}
\label{tab:human_eval}
\end{table}

\paragraph{Human Evaluation}
Despite F1 scores being the most commonly used metrics for evaluating the quality of QA models, there are often multiple ways to answer questions, which makes the automatic metrics less than perfect. Therefore, we use human evaluations to further compare our ChatQA-1.0-70B with GPT-4. In this human evaluation, we ask annotators to verify the facts in ChatQA-1.0-70B and GPT-4's outputs and determine which model provides a more accurate response to the question~\footnote{More human evaluation setup can be found in the Appendix~\ref{appendix:human_eval}.}.  
The human evaluation results over the 10 datasets are shown in Table~\ref{tab:human_eval}. We first find that our ChatQA-1.0-70B and GPT-4 are tie most of the time (69.09\%), and GPT-4 achieves slightly higher win rate ($\sim$3.3\%) than ours. This further confirms our model has powerful capability to produce correct answers. Second, we find that our model achieves slightly better win rate than GPT-4 in ConvFinQA, which indicates the strong arithmetic calculation ability of our model. Third, we find that GPT-4 achieves significantly better win rate in SQA task, which suggests that there is still a gap between our model and GPT-4 on tabular reasoning task.

\begin{table}[t]
\renewcommand{\arraystretch}{1.03}
\centering
\begin{adjustbox}{width={0.8\textwidth},totalheight={\textheight},keepaspectratio}
\begin{tabular}{lccccc}
\toprule
Models                   & Avg-text & Avg-table  & \begin{tabular}[c]{@{}c@{}} Avg-table\\w/o HDial \end{tabular} &  Avg-ret & Avg-nonret \\ \midrule
% ChatQA-1.0-13B     & 45.80   & 62.68      & 40.01   & 61.72      \\
ChatQA-1.0-70B     & 48.88 & 66.42  & 71.42  & 42.33   & 65.96      \\ 
Llama3-ChatQA-1.5-8B   & 47.74 & 72.50 & 75.87 & 41.63 & 68.71  \\
Llama3-ChatQA-1.5-70B   &  \textbf{49.79} & \textbf{77.99} & \textbf{82.85} & \textbf{43.89} & \textbf{72.62}\\
\midrule
GPT-3.5-Turbo-0613 & 46.07   & 60.40  & 66.89 & 40.41   & 60.33      \\
GPT-4-0613         & 46.96   & 70.10 & 80.25 & 41.58   & 66.22      \\ 
GPT-4-Turbo-2024-04-09 & 46.90 & 70.67 & 82.07 & 41.80 & 66.27 \\ \bottomrule
\end{tabular}
\end{adjustbox}
\vspace{.1cm}
\caption{Fine-grained studies on average scores of different dataset types. Avg-text covers datasets where the documents only have text, including Doc2Dial, QuAC, QReCC, CoQA, DoQA, TopiOCQA, and INSCIT. 
Avg-table covers datasets with table in the documents, including ConvFinQA, SQA, and HybriDial.
Avg-ret covers datasets with long documents requiring retrieval, including Doc2Dial, QuAC, QReCC, TopiOCQA, and INSCIT.
Avg-nonret covers datasets with short documents which do not require retrieval, including CoQA, DoQA, ConvFinQA, SQA, and HybriDial.}
\label{tab:finegrained_cqa_results}
\end{table}

\subsection{Fine-grained Analyses}
In Table~\ref{tab:finegrained_cqa_results}, we further compared our models and OpenAI models across different dataset types in our conversational QA benchmarks. 
% In the comparison between ChatQA-1.0-70B and GPT-4-0613, ChatQA-1.0-70B achieves better results in text-only documents (avg-text), demonstrating its superior text understanding ability. Meanwhile, GPT-4-0613 shows better QA capability in tabular data given the comparisons in avg-table.
ChatQA-1.0-70B achieves better results in text-only documents (Avg-text), while GPT-4-0613 shows better QA capability in tabular data given (Avg-table). Regarding the datasets that require or do not require retrieval (Avg-ret and Avg-nonret), ChatQA-1.0-70B and GPT-4-0613 are comparable. 

Furthermore, Llama3-ChatQA-1.5-70B surpasses GPT-4-0613 and GPT-4-Turbo in all categories.

% In the comparison between ChatQA-1.0-13B and GPT-3.5-Turbo, ChatQA-1.0-13B demonstrates better tabular QA capability, given the scores in Avg-table. ChatQA-1.0-13B also shows better scores in datasets where documents do not need retrieval, while being on par with GPT-3.5-Turbo in text-only documents and documents that need retrieval.

\begin{table}[t]
\renewcommand{\arraystretch}{1.03}
\centering
\begin{adjustbox}{width={0.55\textwidth},totalheight={\textheight},keepaspectratio}
\begin{tabular}{lcccc}
\toprule
Models    &   ChatRAG  &    Avg-ret & Avg-nonret \\ \midrule
ChatQA-1.0-70B   &  \textbf{54.14}  & 42.33  & \textbf{65.96}      \\
- w/ ``top-5'' chunks & 54.04  & \textbf{42.91}   &   65.16    \\  \bottomrule
\end{tabular}
\end{adjustbox}
\vspace{.1cm}
\caption{Ablation study on using ``top-5'' retrieved chunk as the context for the stage-2 instruction tuning. We report average scores on \textsc{ChatRAG Bench}, five retrieval datasets (Avg-ret) and five non-retrieval datasets (Avg-nonret).}
\label{tab:retrieved_context_results}
\end{table}

\subsection{Top-\emph{k} Chunks for Stage-2 Instruction Tuning}
\label{subsec:ablation_study_narrative_qa}
\vspace{-.1cm}
For all the datasets used in stage-2 tuning, the context is provided as continuous paragraphs or documents containing the answer. In contrast, the model needs to handle the top-\emph{k} retrieved chunks during inference for long documents.
To address this train/test mismatch, we investigate whether replacing some continuous paragraphs with retrieved top-\emph{k} chunks will enhance the model's robustness.

% In this section, we investigate whether replacing some continuous documents with  retrieved top-k chunks will enhance the model's robustness on datasets with long documents that require retrieval. 

We use NarrativeQA for this study, since each question has a corresponding long document. Initially, we use a summary of the long document as the context, which contains the answer. To incorporate discontinuous contexts, we first cut the long document into 300-word chunks. Then, we use the Dragon retriever to retrieve the top-4 chunks related to the question as additional context. Finally, we consider the retrieved four chunks along with the summary of the long document as the ``top-5'' chunks.
\footnote{Note that, we do not directly use top-5 retrieved chunks for training, because they may not contain the answer. In such cases, fine-tuning the model to generate answer could encourage hallucination.}
We use this reconstructed NarrativeQA to replace the original one for the stage-2 instruction tuning. 

In Table~\ref{tab:retrieved_context_results}, we observe that using ``top-5'' chunks as the context for training leads to improvements on the datasets that require retrieval. But it degrades the performance on non-retrieval datasets. Overall, these two models perform comparable. 
It is because incorporating ``top-5'' retrieved chunks in stage-2 tuning aligns with the inference stage where retrieval is needed, which improves the Avg-ret score. However, mixing continuous and discontinuous documents could make the stage-2 tuning less stable, leading to sub-optimal results on non-retrieval datasets. We believe more future work can be conducted regarding the balance of incorporating continuous context and top-\emph{k} retrieved chunks in stage-2 tuning.

\begin{table}[t]
\renewcommand{\arraystretch}{1.0}
\centering
\begin{adjustbox}{width={0.8\textwidth},totalheight={\textheight},keepaspectratio}
\begin{tabular}{lcccccc}
\toprule
Models & \multicolumn{1}{c}{Avg.} & \multicolumn{1}{c}{Doc2Dial} & \multicolumn{1}{c}{QuAC} & \multicolumn{1}{c}{QReCC} & \multicolumn{1}{c}{TopiOCQA} & \multicolumn{1}{c}{INSCIT} \\ \midrule
ChatQA-1.0-70B         & 42.31                       & 39.19                        & 38.33                    & 48.73                     & 51.30                        & 33.98                      \\
- \# of ctx: top-3         & 41.91                       & 37.20                        & 38.35                    & 48.94                     & 52.78                        & 32.27                      \\
- \# of ctx: top-10        & 40.71                       & 37.06                        & 36.95                    & 47.61                     & 49.40                        & 32.53                      \\
- ctx reverse ordering & 42.48                       & 39.08                        & 38.85                    & 49.63                     & 51.16                        & 33.69                      \\
- ctx swing ordering   & 42.30                       & 39.35                        & 38.09                    & 49.09                     & 50.98                        & 33.99                      \\ 
- ctx random ordering  & 42.01 & 39.32 & 38.28 & 48.79 & 50.13 & 33.51 \\
- Dragon Retrieval      & 40.50                       & 37.92                        & 38.44                    & 47.88                     & 50.39                        & 27.87                      \\ \bottomrule
\end{tabular}
\end{adjustbox}
\vspace{.1cm}
\caption{Ablation studies on input context across datasets that require retrieval. All models use SyntheticConvQA. We study the number of contexts used in inputs (\# of ctx), context ordering (reverse, swing, random), and the use of retrieved context from the original Dragon. In comparison, ChatQA-1.0-70B (default setting) uses ``Dragon + Fine-tune'' to retrieve the top-5 contexts, and arranges them sequentially from the first to the fifth context in top-5.}
\label{tab:ctxs_ablation_study}
\end{table}

\subsection{Ablation Studies on Inference Stage}
% ablation studies on ctxs (using different numbers of ctxs, ordering of ctxs, MT Dragon retrival vs. Dragon retrieval)
In Table~\ref{tab:ctxs_ablation_study}, we show ablation studies on how the number of retrieved context/chunks, context ordering, and different retrievers affect the conversational QA and RAG results. 

First, we find that using more contexts as inputs do not always improve the results. Utilizing top-5 contexts as input yields better results compared to using either top-3 or top-10 contexts. Intuitively, more contexts have higher probability to contain correct answers (better recall scores). As a result, using top-5 contexts achieves better results than using top-3. However, as the number of contexts further increases, the model may suffer from ``lost in the middle'' phenomenon~\citep{liu2023lost} and the difficulty of extracting answers from the provided context could also increase, which leads to inferior results by using top-10 contexts.

Second, we study how using different orderings of top-5 contexts affects the results. We compare sequential ordering (from $1$st context to $5$th context) to reverse ordering (from $5$th to $1$st context), swing ordering (given the ``lost in the middle'' phenomenon, we arrange the most relevant context to appear at the beginning and the end of the input context. Hence, the ordering becomes \{$1$st, $3$rd, $5$th, $4$th, $2$nd\}), and random ordering (random shuffle the top-5 contexts). We find using sequential ordering is comparable to using reverse and swing orderings, and random shuffling is slightly worse. Results indicate that our model excels in extracting the correct answer from lengthy contexts, regardless of the answer's location. This is because, during the ChatQA fine-tuning, the answer's location is randomly distributed within the context.

Third, we observe that when we replace ``Dragon + Fine-tune'' with the original non-finetuned Dragon retriever, the average score drops by 1.81 (from 42.31 to 40.50). In addition, the score drops significantly in INSCIT dataset (from 33.98 to 27.87) due to the large performance gap between the two retrievers (as shown in Table~\ref{tab:retrieval_main}). Basically, when the quality of retrieval improves, it directly enhances the performance of question answering.

\begin{table*}[t]
\renewcommand{\arraystretch}{1.0}
\centering
\begin{adjustbox}{width={1.0\textwidth},totalheight={\textheight},keepaspectratio}
\begin{tabular}{ll|c||ccc|ccc||c}
\toprule
\multicolumn{2}{l|}{Models}                                          & \begin{tabular}[c]{@{}c@{}}Avg\\Both \end{tabular}      & \begin{tabular}[c]{@{}c@{}}Avg\\QuAC\end{tabular}       & \begin{tabular}[c]{@{}c@{}}QuAC\\(no*)\end{tabular} & \begin{tabular}[c]{@{}c@{}}QuAC\\(yes*)\end{tabular} & \begin{tabular}[c]{@{}c@{}}Avg\\DoQA\end{tabular}       & \begin{tabular}[c]{@{}c@{}}DoQA\\(no*)\end{tabular} & \begin{tabular}[c]{@{}c@{}}DoQA\\(yes*)\end{tabular} & ChatRAG        \\ \midrule
\multirow{5}{*}{ChatQA-1.0-70B} & - 1k unanswerable$^\dagger$             & 76.88          & 80.89          & 75.10      & 86.67       & 72.88          & 64.49      & 81.26       & 54.16 \\
                            & - \textbf{1.5k unanswerable}$^\dagger$ & \textbf{77.25} & 80.76 & 77.66      & 83.85       & 73.74 & 68.81      & 78.67       & 54.14 \\
                            & - 2k unanswerable$^\dagger$            & 77.10          & 80.82          & 77.59      & 84.05       & 73.38          & 67.95      & 78.80       & 53.86          \\
                            & - 2.5k unanswerable$^\dagger$          & 75.87          & 78.81          & 73.76      & 83.85       & 72.93          & 66.54      & 79.31       & 53.78          \\
                            & - SyntheticConvQA$^\diamondsuit$       & 69.84          & 72.92          & 55.38      & 90.42       & 66.77          & 45.09      & 88.45       & 54.08          \\ \midrule 
\multicolumn{2}{l|}{Llama3-ChatQA-1.5-8B}  & 75.57 & 79.30 & 63.39 & 95.21 & 71.84 & 62.76 & 80.92 & 55.17  \\ 
\multicolumn{2}{l|}{Llama3-ChatQA-1.5-70B}  & 71.86 & 72.59 & 48.25 & 96.93 & 71.13 & 52.24 & 90.01 & 58.25 \\ \midrule
\multicolumn{2}{l|}{Command R+}    & 68.11 & 69.61 & 41.79 & 97.42 & 66.62 & 46.37  & 86.87 & 50.93           \\
\multicolumn{2}{l|}{GPT-3.5-Turbo-0613}                              & 73.27          & 78.34          & 61.91      & 94.76       & 68.21          & 51.99      & 84.43       & 50.37          \\
\multicolumn{2}{l|}{GPT-4-0613}                                      & \textbf{80.73} & 87.42 & 83.45      & 91.38       & 74.05 & 74.28      & 73.82       & 53.90 \\ 
\multicolumn{2}{l|}{GPT-4-Turbo-2024-04-09}  & 80.47 & 88.73 & 80.42 & 97.03 & 72.21 & 72.28 & 72.13 & 54.03           \\ \bottomrule
\end{tabular}
\end{adjustbox}
% \vspace{-.2cm}
\caption{Accuracies on answerable and unanswerable samples across QuAC and DoQA datasets. Avg-Both is the averaged score between QuAC and DoQA, and ChatRAG is the average score on the \textsc{ChatRAG Bench}. $^\diamondsuit$ denotes that the HumanAnnotatedConvQA is replaced with the SyntheticConvQA. *``no'' denotes the unanswerable samples, while ``yes'' denotes the answerable samples. $^\dagger$We conduct ablation studies in terms of the number of unanswerable samples in HumanAnnotatedConvQA. We use 1.5k unanswerable samples for our final ChatQA models, as it produces both high-quality generation and less hallucination.}
\label{tab:unanswerable}
% \vspace{-.2cm}
\end{table*}

\subsection{Evaluation of Unanswerable Case}
\label{sec:unanswerable}

\subsubsection{Evaluation Setup}
In this section, we study another aspect of the model's  capability, which is to discern if a question can be answered within the provided context. Generating an answer in unanswerable case will lead to hallucination. 
To allow this evaluation, we require the model to indicate it when no answer can be found in the given context.

We use QuAC and DoQA datasets which have such unanswerable cases to evaluate such capability. Specifically, for unanswerable case, we consider the model indicating that the question cannot be answered as correct\footnote{We collect a set of heuristic matching patterns from all generated samples to determine if the model suggests that the question is unanswerable. More details can be found in the Appendix~\ref{appendix:unanswerable_case_evaluation}.}, and as for answerable cases, we consider the model not indicating the question is unanswerable as correct (i.e., the model giving an answer). Note that for answerable cases, we only select the samples where correct context is retrieved. 
% Our model checkpoints (ChatQA-1.0-70B w/ 1.5k unanswerable and ChatQA-1.0-70B w/ SyntheticConvQA) are same ones reported in Table~\ref{tab:cqa_main}. 

In the end, we calculate the average accuracy score of unanswerable and answerable cases as the final metric. 
We consider this average accuracy as a reliable metric since it is in the same spirit of F1 metric which measures the harmonic mean of precision and recall scores.

\subsubsection{Results}
In Table~\ref{tab:unanswerable}, we compare our models with OpenAI models across QuAC and DoQA datasets. 
First, we find that compared to using SyntheticConvQA, using HumanAnnotatedConvQA significantly increase the average accuracies on both QuAC and DoQA. It is because the unanswerable annotations are in higher quality for human-annotated data, which significantly improve the accuracy for unanswerable cases. 
Second, OpenAI models show powerful capability in this task, especially for GPT-4-0613. Compared to them, our best model (ChatQA-1.0-70B) achieved a significantly better average accuracy than GPT-3.5-Turbo, while we still has a slight gap compared to GPT-4-0613 (around 3.5\%). 
Third, 
% we find that models achieving higher accuracy on unanswerable samples tends to get lower accuracy on answerable samples, and vice versa. We speculate that when a model tends to be ``aggressive'' and offer somewhat relevant answers to those unanswerable questions, it will boost the accuracy for answerable cases, but reduces accuracy for unanswerable ones. Conversely, when a model is more ``conservative'' and strictly checks if the question can be answered, it will result in the opposite effects. 
we find that Llama3-ChatQA-1.5-8B/70B generally achieves lower accuracies than ChatQA-1.0-70B. It is because the models tend to give responses based on their own knowledge when the answers cannot be found within the context, which results in a low accuracy on unanswerable cases. 
% ``aggressive'' models tend to offer somewhat relevant answers to those unanswerable questions, which boosts the accuracy for answerable cases, but reduces accuracy for unanswerable ones. 
% Conversely, the relatively ``conservative'' ones tending to strictly check if the question can be answered will result in the opposite effects.

In addition, we conduct ablation studies in terms of the number of unanswerable samples in HumanAnnotatedConvQA. We find that using a small amount of unanswerable samples (e.g., 1.5k) is able to achieve remarkable results on the unanswerable evaluation, as well as the overall \textsc{ChatRAG Bench} score. Interestingly, incorporating more unanswerable samples does not necessarily lead to higher accuracy scores in the unanswerable evaluation.

\begin{table}[t]
\renewcommand{\arraystretch}{1.03}
\centering
\begin{adjustbox}{width={0.65\textwidth},totalheight={\textheight},keepaspectratio}
\begin{tabular}{lcccc}
\toprule
Models                & Average & NQ    & TriviaQA & HotpotQA \\ \midrule
Atlas (11B)~\citep{izacard2023atlas}           & 39.4   & 26.7  & 56.9     & 34.7     \\
Raven (11B)~\citep{huang2023raven}           & -   & 29.6  & 65.7     & -        \\
RECOMP (20B)~\citep{xu2024recomp}          & 42.1   & 37.0  & 59.0     & 30.4     \\
InstructRetro (43B)~\citep{wang2023instructretro}   & -      & 38.9  & 65.6     & -        \\
RePlug (65B)~\citep{shi2023replug}    & 44.5 &  28.8 & 72.6 &  32.0 \\
RA-DIT (65B)~\citep{lin2024radit}          & 50.1   & 35.2  & 75.4     & 39.7     \\
Llama3-Instruct-8B~\citep{llama3}    & 42.5   & 30.9  & 70.7     & 26.0     \\
Llama3-Instruct-70B~\citep{llama3}   & 53.6   & 42.7  & 82.4     & 35.6     \\ \midrule
Llama3-ChatQA-1.5-8B  & 52.3   & 42.4  & 81.0    & 33.5     \\
Llama3-ChatQA-1.5-70B & \textbf{58.7}   & \textbf{47.0}  & \textbf{85.6}    & \textbf{42.2}   \\ \bottomrule
\end{tabular}
\end{adjustbox}
\vspace{.1cm}
\caption{Zero-shot exact match scores on Natural Questions (NQ), TriviaQA, and HotpotQA, which were evaluated using the data split from the KILT Benchmark~\citep{petroni2021kilt}.}
\label{tab:single_turn_QA_rag}
\end{table}

\subsection{Evaluation on Single-Turn QA and RAG Benchmark}
% In addition to \textsc{ChatRAG Bench}, we further evaluate our model on QA datasets: Natural Questions~\citep{kwiatkowski2019natural}, TriviaQA~\citep{joshi2017triviaqa}, and HotpotQA~\citep{yang2018hotpotqa} on KILT benchmark~\citep{petroni2021kilt}. We follow the evaluation setup in Command R+, and we use Dragon~\citep{lin2023train} to retrieve top-n context. 

In addition to \textsc{ChatRAG Bench}, we further evaluate Llama3-ChatQA-1.5 models on knowledge-intensive single-turn QA datasets: Natural Questions (NQ)~\citep{kwiatkowski2019natural}, TriviaQA~\citep{joshi2017triviaqa}, and HotpotQA~\citep{yang2018hotpotqa}, and compare them against frontier RAG models. 
We use the Dragon retriever~\citep{lin2023train} to extract the top-k most relevant contexts, which are then used as the inputs for Llama3-Instruct-8B/70B and Llama3-ChatQA-1.5-8B/70B. We report the best results obtained from using the top-15, top-20, and top-25 contexts.
In Table~\ref{tab:single_turn_QA_rag}, we show that, despite its significantly smaller model size, Llama3-ChatQA-1.5-8B performs better than the state-of-the-art RA-DIT (65B) model. Llama3-ChatQA-1.5-70B remarkably outperforms existing frontier RAG models.

% Llama3-ChatQA-1.5 models still achieve superior performance on single-turn RAG datasets, although they are optimized for conversational QA scenarios. Our models remarkably outperform Llama3-Instruct counterparts, and Llama3-ChatQA-1.5-70B achieve comparable results to other state-of-the-art models (i.e., RA-DIT).

\begin{table}[t]
\renewcommand{\arraystretch}{1.03}
\centering
\begin{adjustbox}{width={1.02\textwidth},totalheight={\textheight},keepaspectratio}
\begin{tabular}{lll}
\toprule
\textbf{Context}       & \textbf{Conversation}         & \textbf{Assistant Responses}     \\ \midrule
\begin{tabular}[c]{@{}l@{}}The process for Board hearings has changed. If you're appealing a VA decision \\ under the new process, keep reading below to learn how to request and prepare \\ for a hearing. \textcolor{blue}{You'll follow the new process if you're : Appealing a VA decision} \\ \textcolor{blue}{you received on or after February 19 , 2019.} Participating in the Rapid Appeals \\ Modernization Program Opting into the new process using VA Form 10182 If \\ you re using VA Form 9 to appeal a VA decision you received before February \\ 19, 2019, go to the Board of Veterans' Appeals BVA Hearing page ... The Board \\of Veterans Appeals schedules hearings in docket order the order in which they \\receive requests ...\end{tabular} & 
\begin{tabular}[c]{@{}l@{}}User: What can I learn at my Board hearing?\\ Assistant: Do you want to learn exactly what to expect?\\ ...\\ User: Is it necessary for me to consider requesting a Board hearing?\\ Assistant: You can request a video conference hearing if you want \\ to provide testimony to a judge about your appeal.\\ User: when will I follow the new process for a Board hearing? \\ \\Assistant:\end{tabular} & 
\begin{tabular}[c]{@{}l@{}}ChatQA-1.0-70B: \textcolor{mygreen}{If you're appealing a VA decision you received on} \\ \textcolor{mygreen}{or after February 19, 2019, you'll follow the new process}.\\ \\ GPT-4-0613: \textcolor{mygreen}{You'll follow the new process for a Board hearing if} \\ \textcolor{mygreen}{you're appealing a VA decision you received on or after February 19,} \\ \textcolor{mygreen}{2019}, participating in the Rapid Appeals Modernization Program, or \\ opting into the new process using VA Form 10182.\\ \\ Ground Truth: You will follow the new process if appealing a VA \\ decision if you received it on or after February 19, 2019.\end{tabular} \\ \midrule

\begin{tabular}[c]{@{}l@{}}title: How did Arya survive this confrontation unscathed?\\ There was no time or intention to turn her into White Walker. Let us look at the \\ scene from S04E04. \textcolor{blue}{A white walker brings a baby holding in his lap and puts it} \\ \textcolor{blue}{on an ice table} (or whatever it is) for the arrival of the Night King. ... When Arya \\ drops the dagger, Night King looks surprised. This shows he doesn't know what \\ would happen next. There was no time to think and turn her into a white walker \\ by touch. There was no planning in this scene like S04E04. ...\end{tabular}            & 
\begin{tabular}[c]{@{}l@{}}User: How did Arya survive this confrontation unscathed?\\ \\ Assistant: Arya was quick enough to react, drop the dagger and \\ do the rest.\\ \\ User: How old is the child that they lay on the altar of ice? \\ \\Assistant:\end{tabular}  &
\begin{tabular}[c]{@{}l@{}}ChatQA-1.0-70B: The child appeared to be \textcolor{mygreen}{a baby}.\\ \\ GPT-4-0613: \textcolor{mygreen}{The text does not provide information on the exact age} \\ \textcolor{mygreen}{of the child} that was laid on the ice altar.\\ \\ Ground Truth: a baby \end{tabular}  \\ \midrule

\begin{tabular}{lllll}
\toprule
\textbf{Massif}  &  \textbf{Type of nature reserve} & \textbf{Preserved area} \\ \midrule
Chornohora  & Carpathian Biosphere &  2476.8 ha \\ 
\textcolor{blue}{Uholka / Wide Meadow}   & Carpathian Biosphere &  \textcolor{blue}{11860 ha} \\ 
Svydovets  & Carpathian Biosphere & 3030.5 ha \\ 
Maramoros    & Carpathian Biosphere & 2243.6 ha \\ 
Rozok  & Presov Preserved & 67.1 ha \\ 
Jasmund & Jasmund National Park & 492.5 ha \\
... \\ \bottomrule
\end{tabular}  &

\begin{tabular}[c]{@{}l@{}}User: which of the forests is a carpathian biosphere reserve? \\ \\Assistant: Chornohora, Uholka / Wide Meadow, Svydovets, \\ Maramoros, Kuziy / Trybushany \\ \\ User: which of those covers more than 3100 ha? \\ \\ Assistant: \end{tabular}  &

\begin{tabular}[c]{@{}l@{}}ChatQA-1.0-70B: \textcolor{mygreen}{Uholka / Wide Meadow}\\ \\ GPT-4-0613: \textcolor{red}{Chornohora}, \textcolor{mygreen}{Uholka / Wide Meadow}, \textcolor{red}{Svydovets}, \\ \textcolor{red}{Maramoros} \\ \\ Ground Truth: Uholka / Wide Meadow \end{tabular}  \\ \midrule
title: Summary of enemy aircraft damaged/destroyed (John B. England) \\
\begin{tabular}{lllll}
\toprule
\textbf{Date}  & \textbf{Location} &  \textbf{No.} & \textbf{Type} & \textbf{Status} \\ \midrule
\textcolor{blue}{24 April 1944}  & Munich, Germany & \textcolor{blue}{3} &  Me-110 & \textcolor{blue}{Destroyed} \\ 
27 May 1944 & Strasbourg, France & 1 & Me-109  & Damaged \\
18 August 1944 & Paris, France & 0.5 & Me-109 & Destroyed \\
\textcolor{blue}{13 September 1944} & Nordhausen, Germany & \textcolor{blue}{2.5}  & Me-109 & \textcolor{blue}{Destroyed} \\
6 October 1944 & Berlin, Germany & 1 & Me-109 & Damaged \\
\textcolor{blue}{6 October 1944} & Berlin, Germany & \textcolor{blue}{2} & Me-109 & \textcolor{blue}{Destroyed} \\
\textcolor{blue}{27 November 1944} & Magdeburg, Germany & \textcolor{blue}{4} & FW-190 & \textcolor{blue}{Destroyed} \\
14 January 1945 & Berlin, Germany & 1 & Me-109 & Destroyed \\
... \\ \bottomrule
\end{tabular}  &\begin{tabular}[c]{@{}l@{}}User: what were all of the dates in which john b. england destroyed or \\ damaged enemy planes? \\ \\Assistant: 24 April 1944, 27 May 1944, 18 August 1944, 13 September \\ 1944, 6 October 1944, 27 November 1944, 14 January 1945 ... \\ \\ User: on which of those dates did john b. england destroy or damage \\ more than one enemy plane? \\ \\ Assistant: \end{tabular}  &
\begin{tabular}[c]{@{}l@{}}ChatQA-1.0-70B: \textcolor{mygreen}{6 October 1944, 13 September 1944, 27 November} \\ \textcolor{mygreen}{1944} \textcolor{red}{(missing 24 April 1944)} \\ \\ GPT-4-0613: \textcolor{mygreen}{24 April 1944, 13 September 1944, 6 October 1944,} \\ \textcolor{mygreen}{27 November 1944} \\ \\ Ground Truth: 24 April 1944, 13 September 1944, 6 October 1944, \\27 November 1944 \end{tabular}  \\ \bottomrule

\end{tabular}
\end{adjustbox}
\vspace{0.1em}
\caption{Case studies on ChatQA-1.0-70B and GPT-4-0613. We use \textcolor{blue}{blue} color to highlight the relevant context for the last user question. And we use \textcolor{mygreen}{green} and \textcolor{red}{red} color to highlight the correct and incorrect parts in the assistant responses.}
\label{tab:case_study}
\end{table}

\subsection{Case Study}
In Table~\ref{tab:case_study}, we show four examples of ChatQA-1.0-70B and GPT-4-0613 outputs. More examples can be found in Appendix~\ref{appendix:case_study}.
The first example (about \textit{Board hearing}) is a simple information seeking problem, and both ChatQA-1.0-70B and GPT-4 answer correctly. In the second example (about \textit{Arya}), the model needs to find implicit information (highlighted in blue) for giving the answer. GPT-4 tends to be safe on providing an answer, and it replies that the context does not provide exact information about the age which is also correct. 

Both the third and fourth examples require the model to have good tabular understanding and reasoning ability. In the third example (about \textit{Massif}), ChatQA-1.0-70B gives correct answer by comparing the size of preserved area against 3100 ha, while GPT-4 fails to do so.
In the fourth example (about \textit{John B. England}), ChatQA-1.0-70B correctly lists three dates but misses one date, while GPT-4 correctly answers the question.

\section{Conclusion}
\label{sec:conclusion}
% In this paper, we % propose a two-stage instruction tuning method and 
% build a family of ChatQA models, varying in model sizes from 7B to 70B. 
In this paper, we build a family of models that surpass GPT-4 on conversational QA and RAG.
We introduce \textsc{ChatRAG Bench}, a collection of 10 conversational QA datasets covering comprehensive evaluations on RAG, table-based reasoning, arithmetic calculations, and unanswerable scenarios. Our ChatQA-1.0-70B model built on Llama2 can slightly outperform GPT-4-0613 and GPT-4-Turbo without using any synthetic data from OpenAI GPT models. Remarkably, Llama3-ChatQA-1.5-70B even surpasses GPT-4-Turbo in all categories of \textsc{ChatRAG Bench}.
In addition, we demonstrate that fine-tuning a single-turn query retriever using our curated conversational QA data performs comparably to the state-of-the-art LLM-based query rewriting model, without incurring extra computational time and potential API costs associated with rewriting.
Furthermore, we show that incorporating a small amount of ``unanswerable'' samples can significantly enhance our model's capability to handle scenarios where answers are unavailable. 
% The unanswerable case evaluation highlights that our best model ChatQA-1.0-70B only has a slight gap compared to GPT-4.
% is able to achieve significantly better accuracy compared to GPT-3.5-Turbo, and it is only slightly worse than GPT-4.

% a conversational QA model that achieves GPT-4 level accuracies. Specifically, we propose a two-stage instruction tuning method that can greatly boost LLM's capability in zero-shot conversational QA.

\bibliography{paper}
\bibliographystyle{neurips_2024}

%%%%%%%%%%%%%%%%%%%%%%%%%%%%%%%%%%%%%%%%%%%%%%%%%%%%%%%%%%%%
\newpage
\appendix

\section{ChatQA Instruction Tuning Templates}
\label{appendix:chatqa_instruction_tuning}
\subsection{Stage-1: Supervised Fine-tuning}

The format template of LLM inputs in stage-1 is as follows:

\begin{verbatim}
System: This is a chat between a user and an artificial intelligence assistant. 
The assistant gives helpful, detailed, and polite answers to the user's questions 
based on the context. The assistant should also indicate when the answer cannot be 
found in the context.

User: {Question 1}

Assistant: {Answer 1}

...

User: {Latest Question}

Assistant:
\end{verbatim}

We use the $\{\texttt{Latest Answer}\}$ from Assistant as the supervision for model outputs.

\subsection{Stage-2: Context-Enhanced Instruction Tuning}
\label{appendix:stage2_prompt}

Based on the stage-1 format template, the LLM inputs in stage-2 adds $\texttt{\{Context for Latest Question\}}$ and $\texttt{\{Instruction\}}$ from \texttt{User}, as depicted below:
\begin{verbatim}
System: This is a chat between a user and an artificial intelligence assistant. 
The assistant gives helpful, detailed, and polite answers to the user's questions 
based on the context. The assistant should also indicate when the answer cannot be 
found in the context.

{Context for Latest Question}

User: {Instruction} + {Question 1}

Assistant: {Answer 1}

...

User: {Latest Question}

Assistant:
\end{verbatim}

Same as Stage-1, we use the $\{\texttt{Latest Answer}\}$ from Assistant as the supervision for model outputs.

As for the $\{\texttt{Instruction}\}$, we use different instructions for different datasets based on the answer types. Details can be found below:

``\texttt{Please give a full and complete answer for the question.}'' This is for datasets with long answers. We use it for the HumanAnnotatedConvQA or SyntheticConvQA.
% and a proprietary conversational QA dataset.

``\texttt{Answer the following question with a short span. The answer needs to be just in a few words.}'' This is for datasets with short answers. We use it for SQuAD1.1, SQuAD2.0, NarrativeQA, DROP, ROPES, NewsQA, and Quoref.

``\texttt{Answer the following question with a number from context or the math arithmetic using +, -, *, or /.}'' This is for datasets using arithmetic calculation or extracting numbers from the context. We use it for the TAT-QA dataset where questions need the arithmetic calculation or extracting numbers from the context.

``\texttt{Answer the following question with a short span, or a full and complete answer.}'' This is for datasets with both short and long answers. The model needs to tell whether to generate short or long answers based on the questions. We use it for the TAT-QA dataset where questions do not need the arithmetic calculation, since we observe the lengths of the corresponding answers can be both short and long.

For the stage-1 SFT data used in stage-2 instruction tuning, we keep the same format template as in stage-1. In other words, the $\texttt{\{Context for Latest Question\}}$ and $\texttt{\{Instruction\}}$ are empty.

\section{Synthetic Data Generation}
\textbf{We use a two-step prompting on GPT-3.5-Turbo for obtaining a high-quality multi-turn QA conversation.} In step-1, we prompt GPT-3.5-Turbo to generate a multi-turn QA conversation based on a given document. However, we find that the generated conversation lacks pronouns or common nouns to refer back to words or phrases in dialogue history. Hence, in step-2, we prompt GPT-3.5-Turbo to revise the generated conversation by adding pronouns or common nouns. For both stages, we give GPT-3.5-Turbo few-shot examples to ensure a high-quality generation.

The in-context few-shot prompt in step-1 is as follows (aims to generate multi-turn QA conversation):
\begin{verbatim}
You are a helpful, pattern-following assistant.

role-user:
PersonA has access to information in the following passage:
###There are certain important aspects of college preparation. The student preparing 
for college should define their goals and interests. The student should also plan 
financially and academically. We have a checklist that helps students get ready 
financially and academically.###
Generate a conversation of around four turns between PersonA and PersonU where 
PersonU needs information from PersonA that is in the passage. Use only information 
found in the passage.
PersonU starts the conversation and only asks questions. Most of PersonU's questions 
are follow-up or continuations of previous questions.
Conversation:

role-assistant:
PersonU: I need to know how to prepare for college.
PersonA: You should first define your goals and interest and be sure to know the 
costs and how to plan financially and academically for college.
PersonU: Is there something I can use as a guide to get ready for it?
PersonA: you can use our checklist to get ready financially and academically.
PersonU: Any other tip?
PersonA: Be sure to check out the reasons to attend a College or Career schools.
PersonU: Ok, that sounds good. Can you give me more tips?
PersonA: Another thing to take into consideration is to explore your career options.

role-user:
PersonA has access to information in the following passage:
###There are several disability compensation benefits. One of the benefits is the 
annual clothing allowance. This allowance covers the cost of purchasing new clothes 
up to a limit of 200 dollars annually.###
Generate a conversation of around two turns between PersonA and PersonU where 
PersonU needs information from PersonA that is in the passage. Use only information 
found in the passage.
PersonU starts the conversation and only asks questions. Most of PersonU's questions 
are follow-up or continuations of previous questions.
Conversation:

role-assistant:
PersonU: I heard about a disability benefit for clothing. What is it about?
PersonA: This is a disability compensation benefit known as an annual clothing 
allowance.
PersonU: I am quite interested about such benefits.
PersonA: Would you like to find out if you are eligible?

role-user:
PersonA has access to information in the following passage:
###Vascepa (icosapent ethyl) is a prescription fish oil medication approved for 
treating adults with extremely high triglycerides, defined as a triglyceride level 
of 500 mg/dL or higher. While using this medicine, you may need frequent blood tests 
at your doctor's office. Vascepa is only part of a treatment program that may also 
include diet, exercise, and weight control. Follow your doctor's instructions very 
closely. You should not use Vascepa if you are allergic to icosapent. To make sure 
Vascepa is safe for you, tell your doctor if you have: 1 liver disease; 2 diabetes;
3 a thyroid disorder; 4 problems with your pancreas; 5 a bleeding or blood-clotting 
disorder; 6 if you are allergic to fish or shellfish; or. 7 if you drink large 
amounts of alcohol, Vascepa (icosapent) works in the liver and bloodstream to reduce 
very low-density triglycerides.###
Generate a conversation of around three turns between PersonA and PersonU where 
PersonU needs information from PersonA that is in the passage. Use only information 
found in the passage.
PersonU starts the conversation and only asks questions. Most of PersonU's questions 
are follow-up or continuations of previous questions.
Conversation:

role-assistant:
PersonU: I heard about a medication called Vascepa. What is it used for?
PersonA: Vascepa is a prescription fish oil medication used for treating adults with 
extremely high triglyceride levels.
PersonU: Are there any side effects or precautions I should be aware of?
PersonA: While using Vascepa, you may need frequent blood tests, and it's important 
to follow your doctor's instructions closely. It's important to talk to your doctor 
about any allergies or existing conditions you have.
PersonU: Is it effective in lowering cholesterol levels?
PersonA: Vascepa is used together with a low-fat diet and other treatments to lower 
total cholesterol in people with severe high cholesterol.

role-user:
PersonA has access to information in the following passage:

{Document}

Generate a conversation of around three turns between PersonA and PersonU where 
PersonU needs information from PersonA that is in the passage. Use only information 
found in the passage.
PersonU starts the conversation and only asks questions. Most of PersonU's questions 
are follow-up or continuations of previous questions.
Conversation:

role-assistant:
\end{verbatim}

The in-context few-shot prompt in step-2 is as follows (aims to revise the multi-turn QA conversation by adding pronouns and common nouns):

\begin{verbatim}
You are a helpful, pattern-following assistant.

role-user:
Given the following conversation between PersonU and PersonA:
###PersonU: I've heard about adobo sauce in Mexican cuisine. Can you tell me more 
about it?
PersonA: Adobo sauce is a dark red sauce made from ground chiles, herbs, and 
vinegar. It's used as a marinade and cooking sauce in Mexican and Southwestern 
cuisines.
PersonU: What are the main ingredients in adobo sauce?
PersonA: The main ingredients are typically chile peppers, garlic, and vinegar. 
However, there can be variations in the recipe.
PersonU: How can I make adobo sauce at home?
PersonA: One way to make adobo sauce at home is by using smoked jalapeno peppers 
called chipotles. You can stew the chipotles in a sauce with tomatoes, garlic, 
vinegar, salt, and spices. It's also possible to find canned adobo sauce in the 
Hispanic foods section of many grocery stores.
PersonU: Are there any other uses for adobo sauce?
PersonA: Adobo sauce can be used as a condiment or as a marinade for various dishes. 
It enhances the flavor of meats and can be used as a cooking and serving sauce. 
Additionally, adobo is also used in a stew-like presentation of meat prepared in 
this sauce.
PersonU: Is adobo sauce easy to preserve?
PersonA: Yes, you can preserve adobo sauce by freezing or canning it when you're 
done making it. This allows you to store it for up to a year and enjoy its flavors 
whenever you want.###
Modify PersonU's questions to use pronouns or common nouns to refer back to words 
or phrases in PersonU's previous questions or PersonA's previous responses.

role-assistant:
PersonU: I've heard about adobo sauce in Mexican cuisine. Can you provide more 
information about it?
PersonA: Adobo sauce is a dark red sauce made from ground chiles, herbs, and vinegar. 
It's used as a marinade and cooking sauce in Mexican and Southwestern cuisines.
PersonU: What are the main ingredients in the sauce?
PersonA: The main ingredients are typically chile peppers, garlic, and vinegar. 
However, there can be variations in the recipe.
PersonU: How can I make the sauce at home?
PersonA: One way to make it at home is by using smoked jalapeno peppers called 
chipotles. You can stew them in a sauce with tomatoes, garlic, vinegar, salt, and 
spices. It's also possible to find canned adobo sauce in the Hispanic foods section 
of many grocery stores.
PersonU: Are there any other uses for it?
PersonA: It can be used as a condiment or as a marinade for various dishes. It 
enhances the flavor of meats and can be used as a cooking and serving sauce. 
Additionally, it is also used in a stew-like presentation of meat prepared in 
this sauce.
PersonU: Is it easy to preserve?
PersonA: Yes, you can preserve it by freezing or canning it when you're done 
making it. This allows you to store it for up to a year and enjoy its flavors 
whenever you want.

role-user:
Given the following conversation between PersonU and PersonA:
###PersonU: I've heard about a test called urinalysis. What is it used for?
PersonA: Urinalysis is a test that evaluates a sample of your urine. It is used to 
detect and assess a wide range of disorders, such as urinary tract infection, 
kidney disease, and diabetes. It examines the appearance, concentration, and content 
of urine.
PersonU: Can it help diagnose bladder cancer?
PersonA: Yes, urinalysis can be used as a screening and diagnostic tool for 
bladder cancer. One way to test for bladder cancer is to check for blood in the 
urine, which is called hematuria. This can be done during a urinalysis by examining 
the urine cells under a special microscope.
PersonU: What other conditions or diseases can be detected through urinalysis?
PersonA: Urinalysis is commonly used to diagnose urinary tract or kidney infections, 
evaluate causes of kidney failure, screen for the progression of chronic conditions 
like diabetes mellitus and high blood pressure, and it can be used in combination 
with other tests to diagnose certain diseases. It is a versatile test that helps 
detect abnormalities in the urinary tract.
PersonU: Where can I get a urinalysis done?
PersonA: Urinalysis can be performed in various healthcare settings, including 
doctors' offices, urgent care facilities, laboratories, and hospitals. You can 
provide a urine sample in a specimen cup, and the sample can be analyzed on-site 
or sent to a laboratory for testing.
PersonU: How much urine is typically needed for a urinalysis?
PersonA: Usually, only small amounts of urine, ranging from 30 to 60 ml, are required 
for urinalysis testing. It's a non-invasive and relatively simple test that provides 
valuable information about your health.###
Modify PersonU's questions to use pronouns or common nouns to refer back to words 
or phrases in PersonU's previous questions or PersonA's previous responses.

role-assistant:
PersonU: I've heard about a test called urinalysis. What is it used for?
PersonA: Urinalysis is a test that evaluates a sample of your urine. It's used to 
detect and assess a wide range of disorders, such as urinary tract infection, 
kidney disease, and diabetes. It examines the appearance, concentration, and 
content of urine.
PersonU: Can it help diagnose bladder cancer?
PersonA: Yes, urinalysis can be used as a screening and diagnostic tool for bladder 
cancer. One way to test for bladder cancer is to check for blood in the urine, which 
is called hematuria. This can be done during a urinalysis by examining the urine 
cells under a special microscope.
PersonU: What other conditions or diseases can be detected through this test?
PersonA: Urinalysis is commonly used to diagnose urinary tract or kidney infections, 
evaluate causes of kidney failure, screen for the progression of chronic conditions 
like diabetes mellitus and high blood pressure, and it can be used in combination 
with other tests to diagnose certain diseases. It is a versatile test that helps 
detect abnormalities in the urinary tract.
PersonU: Where can I go to get this test done?
PersonA: Urinalysis can be performed in various healthcare settings, including 
doctors' offices, urgent care facilities, laboratories, and hospitals. You can 
provide a urine sample in a specimen cup, and the sample can be analyzed on-site 
or sent to a laboratory for testing.
PersonU: How much urine is typically needed for the test?
PersonA: Usually, only small amounts of urine, ranging from 30 to 60 ml, are 
required for urinalysis testing. It's a non-invasive and relatively simple test 
that provides valuable information about your health.

role-user:
Given the following conversation between PersonU and PersonA:

{multi-turn QA conversation}

Modify PersonU's questions to use pronouns or common nouns to refer back to words 
or phrases in PersonU's previous questions or PersonA's previous responses.

role-assistant:


\end{verbatim}

\section{More Details and Results for Retrieval in Conversational QA}
\label{appendix:multiturn_qa_retrieval}

\subsection{Query Rewriting Prompts for GPT-3.5-Turbo}
\label{appendix:query_rewriting_prompts_GPT-3.5-Turbo}
The in-context few-shot query rewriting prompt we use for GPT-3.5-Turbo is as follows:

\begin{verbatim}
You are a helpful, pattern-following assistant.

role-user:
Given the following conversation between PersonU and PersonA:
PersonU: Hello, I would like to know what to do if I do not agree with any decision.
PersonA: disagree with our decision about your monthly income adjustment amounts?
PersonU: no. Where can I find my SHIP contact information?
PersonA: You can find your local SHIP contact information in the back of your 
Medicare & You 2020 Handbook online.
PersonU: and how do they calculate the adjustments?
Instead of having this entire conversation, how can PersonU get what he or she is 
looking for using a single question? Respond with that question.

role-assistant:
How is the calculation for adjustments made by SHIP determined?

role-user:
Given the following conversation between PersonU and PersonA:
PersonU: I need to know how to prepare for college.
PersonA: You should first define your goals and interest and be sure to know the 
costs and how to plan financially and academically for college.
PersonU: Is there something I can use as a guide to get ready for it?
Instead of having this entire conversation, how can PersonU get what he or she is 
looking for using a single question? Respond with that question.

role-assistant:
What resources or guides can I use to help me prepare for college?

role-user:
Given the following conversation between PersonU and PersonA:

{Dialogue History + Latest Question}

Instead of having this entire conversation, how can PersonU get what he or she is 
looking for using a single question? Respond with that question.

role-assistant:

\end{verbatim}

\subsection{More Results for Retrieval in Conversational QA}
\label{appendix:More_results_for_Multi-Turn QA Retrieval}

\begin{table*}[h]
\renewcommand{\arraystretch}{1.03}
\centering
\begin{adjustbox}{width={0.99\textwidth},totalheight={\textheight},keepaspectratio}
\begin{tabular}{lcccccccccccc}
\toprule
\multirow{2}{*}{Models}                                                                                                         & \multicolumn{2}{c}{Average} & \multicolumn{2}{c}{Doc2Dial} & \multicolumn{2}{c}{QuAC} & \multicolumn{2}{c}{QReCC} & \multicolumn{2}{c}{TopiOCQA} & \multicolumn{2}{c}{INSCIT} \\ \cmidrule(lr){2-3} \cmidrule(lr){4-5} \cmidrule(lr){6-7} \cmidrule(lr){8-9} \cmidrule(lr){10-11} \cmidrule(lr){12-13}
& top-1         & top-5        & top-1       & top-5      & top-1       & top-5   & top-1        & top-5     & top-5*       & top-20*       & top-5*        & top-20*        \\ \midrule
Dragon (w/ dialog history)   & 46.29       & 73.09    & 43.33         & 75.61        & 56.8        & 82.86      & 46.17       & 81.96       & 57.68        & 78.80   & 27.49 & 46.22           \\
Dragon + Rewrite (w/ dialog history)      & 47.57        & 74.12     & 44.54         & 76.98        & 57.23       & 83.04      & 46.45       & 82.60        & 60.94        & 81.74 & 28.69 & 46.22            \\ 
Dragon + Rewrite (w/ single query only)     & \textbf{54.46}       & \textbf{80.13}   & 47.60          & 80.60         & 47.10        & 77.15      & 51.73       & 85.78       & 73.07        & 88.19  & 52.79 & 68.92           \\ \midrule
Dragon + Fine-tune (w/ dialog history) &  52.72 & 80.67 & 48.94 & 83.01 & 52.64 & 81.95 & 50.73 & 87.17 & 67.86 & 86.28 & 43.43 & 64.94 \\
Dragon + Fine-tune + Rewrite (w/ dialog hisotry)    & \textbf{53.17}      & \textbf{80.84}   & 49.30          & 84.64        & 55.04       & 83.23      & 51.23       & 87.99       & 60.50         & 81.03  & 49.80 & 67.33        \\
\bottomrule
\end{tabular}
\end{adjustbox}
\caption{Comprehensive multi-turn retrieval results across five datasets.}
\label{tab:retrieval_all}
\end{table*}

In Table~\ref{tab:retrieval_all}, we show comprehensive comparisons between fine-tuning and rewriting methods.

Interestingly, we find that compared to only using the rewritten query as input (Dragon + Rewrite (w/ single query only)), giving additional dialog history (Dragon + Rewrite (w/ dialog history)) makes the average scores significantly drop. This is because Dragon is originally pretrained on single-turn queries, it will naturally have better generalization ability when a single-turn rewritten query instead of a multi-turn conversation is provided. And the rewritten query already contains sufficient information from the dialog history. 

In addition, we observe that ``Dragon + Fine-tune'' performs on par with ``Dragon + Fine-tune + Rewrite''. In other words, for the multi-turn fine-tuning method, replacing original query with rewritten query as inputs yields comparable results. This is because rewritten query will not provide much additional information for the model since the dialog history has already been provided, and it might even cause negative effects (e.g., results on the TopiOCQA dataset) since it makes the whole conversation input not natural. This once again demonstrates how effective the fine-tuning method is in equipping the model with the capability to comprehend the multi-turn context.

\section{Details of \textsc{ChatRAG Bench}}
\label{appendix:benchmarks}

\subsection{Data Statistics}
% dev or test set
% number of dialogs and number of turns
% some datasets have unanswerable cases but too few examples, hence we discard them in unanswerable case evaluation.

\paragraph{Doc2Dial}
We use the test set of Doc2Dial for the evaluation. It consists of 719 dialogs with 3939 user-agent turns.

\paragraph{QuAC} 
We use the validation set of QuAC for the evaluation since its test set cannot be directly obtained. Its validation set consists of 1000 dialogs with 7354 user-agent turns. Among these 7354 user-agent turns, there are 1486 (around 20.2\%) unanswerable questions. For the combined evaluation of answerable and unanswerable questions, we set the ground truth response for the unanswerable question as ``\texttt{Sorry. I cannot find the answer based on the context.}'', same as the one we set in our stage-2 tuning. For the fair comparison, we replace the unanswerable responses in the baseline models (i.e., Llama2-SFT/Chat, GPT-3.5-Turbo, GPT-4) with the same sentence (Details can be found in Appendix~\ref{appendix:unanswerable_case_evaluation}).

\paragraph{QReCC}
We use the test set of QReCC for the evaluation. Its test set includes some dialogue samples source from the QuAC dataset. To avoid overlap with other benchmark datasets, we remove those QuAC-sourced samples, resulting in 2805 user-agent turns.

\paragraph{TopiOCQA}
We use the validation set of TopiOCQA since its test set is not available yet. Its validation set consists of 205 dialogs with 2514 user-agent turns. Each question has a total of four human-written answers. It has unanswerable cases. However, we do not include this dataset in unanswerable case evaluation because we only find 46 questions with no answer being found, which is not sufficient enough for the unanswerable case evaluation. Given that this dataset use the whole Wikipedia as the retrieval corpus, we leverage its topic information for multi-turn retrieval since we focused more on conversational QA over documents which typically narrow down to a specific topic or domain.

\paragraph{INSCIT}
We use the validation set of TopiOCQA since its test set is not available yet. Its validation set consists of 86 dialogues with 502 user-agent turns. Each question has an average of 1.9 human-written answers. It has unanswerable response for a few questions. However, we do not include this dataset in unanswerable case evaluation because every question always has at least one answerable response, which means every question is answerable. INSCIT also use the whole Wikipedia as the retrieval corpus. Hence, we also leverage its topic information for multi-turn retrieval for the same reason mentioned in TopiOCQA.

\paragraph{CoQA}
We use the validation set of CoQA since its test set cannot be directly obtained. Its validation set consists of 500 dialogues with 7983 user-agent turns. Each question has a total of four human-written answers. It has unanswerable cases. However, we do not include this dataset in unanswerable case evaluation for the same reason as in the TopiOCQA dataset. We only find 13 questions with no answer being found, which is not sufficient enough for the unanswerable case evaluation.

\paragraph{DoQA}
We use the test set of DoQA for the evaluation. Its test set consists of 1200 dialogues with 5394 user-agent turns across Cooking, Travel and Movie domains. Among 5394 user-agent turns, there are 1479 (around 27.4\%) unanswerable questions. We use the same strategy as the one mentioned in QuAC dataset to incorporate unanswerable samples into the evaluation.

\paragraph{ConvFinQA}
We use the validation set of ConvFinQA since its test set cannot be directly obtained. Its validation set consists of 421 dialogues with 1490 user-agent turns. 

\paragraph{SQA}
We use the test set of SQA for the evaluation. Its test set consists of 1025 dialogues with 3100 user-agent turns.

\paragraph{HybriDial}
We use the test set of HybriDial for the evaluation. Its test set consists of 243 dialogues with 1111 user-agent turns.

\subsection{Details of Incorporating Table as Context}
We use the markdown format to process the table in the document (i.e., using ``$\mid$'' to separate each cell in the table).
However, as the number of rows in a table increases, it is harder for the model to link the values of cells in the last few rows to their corresponding headers in the table. To address this issue, we append the headers to the corresponding cell values in each row, so as to make it easier for the model to capture the connections between the cells and their respective headers. For example, a original markdown table

\begin{verbatim}
| Boat | Crew | Nation | Date | Meet | Location | 
| M1x | Mahe | FR | 2009 | | Poznan |
| M2- | Hamish | US | 2012 | Olympics | Lucerne |
| M2+ | Igor | DE | 1994 | | Indianapolis |
\end{verbatim}

will be converted to
\begin{verbatim}
| M1x (Boat) | Crew: Mahe | Nation: FR | Date: 2009 | Meet: | Location: Poznan |
| M2- (Boat) | Crew: Hamish | Nation: US | Date: 2012 | Meet: Olympics | Location: Paris |
| M2+ (Boat) | Crew: Igor | Nation: DE | Date: 1994 | Meet: | Location: Indianapolis |
\end{verbatim}

We apply this table processing strategy to all tabular datasets. 
% including ConvFinQA, SQA, HybriDial, and TAT-QA.

\subsection{Prompts for the Benchmarks}
We adopt the prompt format mentioned in \S\ref{appendix:stage2_prompt} for benchmarking ChatQA, Llama2-Chat, Llama3-Instruct, GPT-3.5-Turbo, GPT-4, and GPT-4-Turbo. For Llama2-Chat and Llama3-Instruct, we modify the prompt based on their own templates, and we convert the prompt into the chat completion template for OpenAI models.
We keep the \texttt{\{Context for Latest Question\}} the same for all models. We adjust the \texttt{\{Instruction\}} for the baselines (i.e., Llama2-Chat, GPT-3.5-Turbo and GPT-4) to ensure they do not get sub-optimal results. 
We use different instructions for test benchmarks with different answer types (e.g., long answer, short answer, arithmetic calculation). The \texttt{\{Instruction\}} for all models on the benchmarks are as follows:

\subsubsection{ChatQA}
We keep the \texttt{\{Instruction\}} consistent between the data blends in stage-2 and test benchmarks based on different answer types. 

We use ``\texttt{Please give a full and complete answer for the question.}'' for DoQA, INSCIT, HybriDial, Doc2Dial, QuAC, and QReCC, since these datasets generally have long answers for the questions.

We use ``\texttt{Answer the following question with a short span, or a full and complete answer.}'' for SQA and TopiOCQA, since these datasets have both short and long answers based on the questions.

We use ``\texttt{Answer the following question with a short span. The answer needs to be just in a few words.}'' for CoQA, since it generally has short answers for the questions.

We use ``\texttt{Answer the following question with a number from context or the math arithmetic using +, -, *, or /.}'' for ConvFinQA, since this dataset requires the model to either extract numbers from the context or do arithmetic calculation. We will calculate the number based on the arithmetic formula the model generates and compare it with the gold answer.

\subsubsection{Llama2-Chat \& Llama3-Instruct}
Note that we tried original Llama2-Chat prompt template\footnote{\url{https://gpus.llm-utils.org/llama-2-prompt-template/}}, which gives slightly worse results compared to using the one in \S\ref{appendix:stage2_prompt}. We have tried several \texttt{\{Instruction\}} for Llama2-Chat. We find the ones below works the best.

We use ``\texttt{Please give an answer in just one sentence.}'' for DoQA, INSCIT, HybriDial, Doc2Dial, QuAC, and QReCC, since these datasets generally have long answers but within one sentence. We notice that it is important to give Llama2-Chat specific instruction like ``one sentence'' instead of ``full and complete answer'' to prevent the model from generating very long answers.

We use ``\texttt{Answer the following question with a short span, or one sentence.}'' for TopiOCQA, since this dataset has both short and long answers based on the questions, and the long answers are generally within one sentence.

We use ``\texttt{Answer the following questions with one or a list of entities.}'' for SQA, since the answer for this dataset always consists of one or a list of entities from the context.

We use ``\texttt{Answer the following question with a short span. The answer needs to be just in a few words.}'' for CoQA, since it generally has short answers for the questions.

We use ``\texttt{Answer the following question with just a number from context or just the math arithmetic using +, -, *, or /.}'' for ConvFinQA, since this dataset requires the model to either extract numbers from the context or do arithmetic calculation. We extract the arithmetic formula generated by the model and use a calculator to get its final result.

\subsubsection{GPT-3.5-Turbo \& GPT-4 \& GPT-4-Turbo}
We have tried several \texttt{\{Instruction\}} for GPT-3.5-Turbo, GPT-4, and GPT-4-Turbo, we find the ones below works the best.

We use ``\texttt{Please give an answer in just one sentence.}'' for DoQA, INSCIT, HybriDial, Doc2Dial, QuAC, and QReCC, since these datasets generally have long answers but within one sentence. Similar to Llama2-Chat, we also find OpenAI models tend to generate quite long answers given the instruction of ``full and complete answer''. Hence, we make the instruction more specific (i.e., ``one sentence'') to prevent the model from generating very long answers.

We use ``\texttt{Answer the following questions in JUST a few words or one sentence.}'' for TopiOCQA, since this dataset has both short and long answers based on the questions, and the long answers are generally within one sentence.

We use ``\texttt{Answer the following questions with one or a list of entities. Do not give a detailed explanation. Answer needs to be as short as possible.}'' for SQA. We find that OpenAI models often give detailed explanation for SQA dataset unless specifically instructed not to. 

We use ``\texttt{Answer the following question with a short span. The answer needs to be just in a few words.}'' for CoQA, since it generally has short answers for the questions.

We use ``\texttt{Answer the following questions with just a number from context or just the math arithmetic using +, -, *, or /.}'' for ConvFinQA, since this dataset requires the model to either extract numbers from the context or do arithmetic calculation. We extract the arithmetic formula generated by the model and use a calculator to get its final result.

\vspace{-1.5em}
\section{Human Evaluation}
\label{appendix:human_eval}
We conduct human evaluation across the ten test benchmark datasets. We randomly select 60 samples for each dataset, and each sample is labelled by three annotators, which results in a total of 1800 annotations.

\begin{figure}[t]
    \centering
    \includegraphics[width=0.98\linewidth]{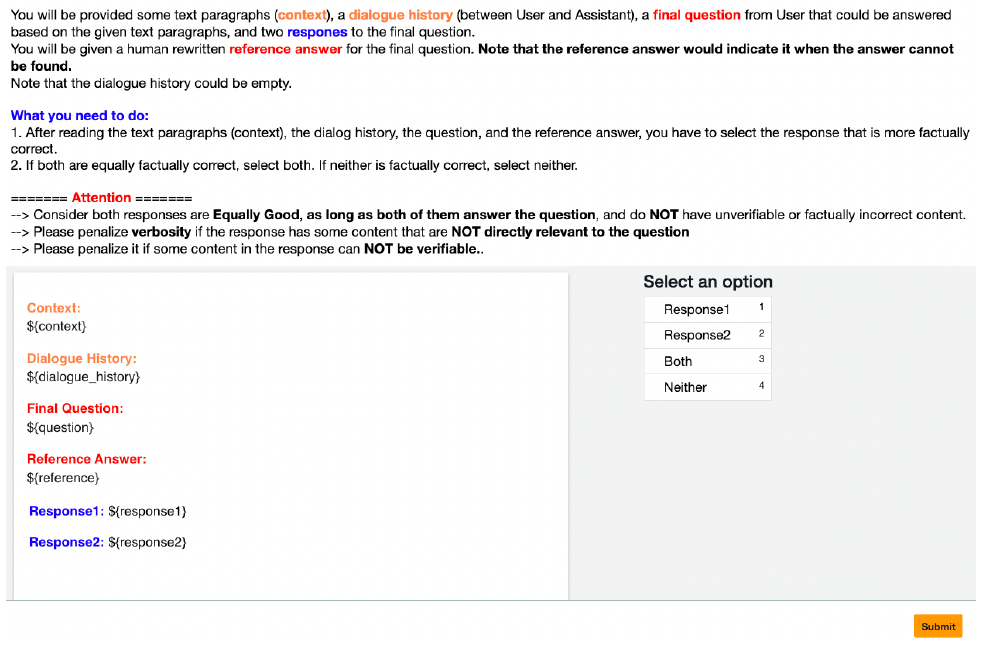}
    \vspace{-2em}
    \caption{Human evaluation instructions and the interface for annotators.}
    \label{fig:human_eval_instrcution}
\end{figure}

We ask the annotators to verify the facts in models' outputs and determine which model provides a more accurate response to the question. We use Amazon Mechanical Turk platform to conduct the human evaluation. The details of our human evaluation instructions and the interface shown to annotators are illustrated in Figure~\ref{fig:human_eval_instrcution}.

\section{Unanswerable Case Evaluation}
\label{appendix:unanswerable_case_evaluation}
For unanswerable case, we consider the model indicating that the question cannot be answered as correct. For answerable cases, we consider the model not indicating the question is unanswerable as correct (i.e., the model giving an answer).
We find that ChatQA always generates ``\texttt{Sorry. I cannot find the answer based on the context}'' when it finds the question is unanswerable, since we use this sentence as a response for unanswerable questions in the stage-2 tuning. As for Llama2-Chat, GPT-3.5-Turbo, and GPT-4, their outputs for unanswerable questions usually follow some specific patterns. Below, we list the heuristic matching patterns derived from all generated samples in DoQA and QuAC datasets to determine if the model suggests that the question is unanswerable. When we find the generated output contains the pattern listed below, we interpret it as an indication that the question is unanswerable, and we will replace its answer with ``\texttt{Sorry. I cannot find the answer based on the context.}'' for the F1 evaluation.

% \begin{verbatim}
% i'm not sure, cannot find, does not provide, cannot provide, cannot answer, 
% cannot be found, cannot be determined, don't have information, do not have 
% information, couldn't find, no information in the context, does not mention,
% not explicitly mentioned, i don't have any, i do not have any, does not 
% specify, doesn't provide, not able to, unable to, doesn't specify, there is 
% no information, there is no mention, not mentioned, i don't have enough 
% information, there is no specific information, there is no specific mention,
% no information found, I don't have that information 
% \end{verbatim}
\begin{verbatim}
cannot find, can't find, not able to, unable to, does not provide, cannot 
provide, cannot answer, couldnot answer, can't answer, couldn't answer, 
cannot be found, cannot be determined, do not have, couldn't find, no 
information, does not mention, doesn't mention, not explicitly mentioned, 
not explicitly explain, can not find, could not find, does not specify, 
doesn't provide, doesn't specify, there is no, not mentioned, don't have, 
don't know, does not include, doesn't include, does not contain, doesn't 
contain, not provided, does not indicate, doesn't indicate, does not 
disclose, doesn't disclose
\end{verbatim}

\section{Case Study}
\label{appendix:case_study}

More examples on ChatQA-1.0-13B, ChatQA-1.0-70B, GPT-3.5-Turbo-0613, and GPT-4-0613 can be found in Table~\ref{tab:case_study_more}. We use \textcolor{blue}{blue} color to highlight the relevant context for the last user question. And we use \textcolor{mygreen}{green} and \textcolor{red}{red} color to highlight the correct and incorrect parts in the assistant responses.

\vspace{1cm}

\begin{table}[h]
\renewcommand{\arraystretch}{1.03}
\centering
\begin{adjustbox}{width={0.99\textwidth},totalheight={\textheight},keepaspectratio}
\begin{tabular}{lll}
\toprule
\textbf{Context}       & \textbf{Conversation}         & \textbf{Assistant Responses}     \\ \midrule

\begin{tabular}{lll}
\toprule
\textbf{Pos}  &  \textbf{Rider} & \textbf{Time/Retired} \\ \midrule
1  & Loris Capirossi &  38:04.730 \\
2  & Valentino Rossi & +0.180 \\
3  & Jeremy McWilliams & +0.534 \\
4  & Tohru Ukawa & +0.537 \\
5  & Shinya Nakano & +0.742 \\
6 & Ralf Waldmann & +7.019 \\
7 & Franco Battaini & +20.889 \\
8 & Stefano Perugini & +20.891 \\
9 & Jason Vincent & +21.310 \\
10 & Anthony West & +26.816 \\
11 & Alex Hofmann & +26.933 \\
... \\
23 & Arno Visscher & +1:40.635 \\
24 & Henk Van De Lagemaat & \textcolor{blue}{+1 Lap} \\
Ret & Maurice Bolwerk & Retirement \\
Ret & Andre Romein & Retirement \\
Ret & Roberto Rolfo & Retirement \\
Ret & Marcellino Lucchi & Retirement \\ \bottomrule
\end{tabular}  &

\begin{tabular}[c]{@{}l@{}}User: what racers are shown? \\ \\Assistant: Loris Capirossi, Valentino Rossi, \\ Jeremy McWilliams, Tohru Ukawa, ... , \\ Henk Van De Lagemaat, Maurice Bolwerk, \\ Roberto Rolfo, Marcellino Lucchi. \\ \\ User: of those, what times were recorded? \\ \\ Assistant: 38:04.730, +0.180, +0.534, \\ +0.537, ... , +1:40.635, +1 Lap, Retirement. \\ \\ User: of those times, which were not \\ actual times? \\ \\ Assistant: +1 Lap, Retirement \\ \\ User: which of was an actual finish? \\ \\ Assistant: \end{tabular}  &

\begin{tabular}[c]{@{}l@{}} ChatQA-1.0-13B: \textcolor{red}{Loris Capirossi} \\ \\ ChatQA-1.0-70B: \textcolor{mygreen}{+1 Lap}\\ \\ GPT-3.5-Turbo-0613: \textcolor{red}{38:04.730} \\ \\ GPT-4-0613: \textcolor{red}{38:04.730} \\ \\ Ground Truth: +1 Lap \end{tabular}  \\ \midrule

\begin{tabular}{lll}
\multicolumn{3}{l}{\begin{tabular}[c]{@{}l@{}}2014 compared to 2013 mst 2019s net sales  \\ decreased \$ 305 million, or 3\% in 2014 as \\ compared to 2013 ... space systems 2019 \\ operating results included the following \\ (in millions):\end{tabular}} \\
\toprule
  &  \textbf{net sales} & \textbf{backlog at year-end} \\ \midrule
2015  & \$ 9105 &  \textcolor{blue}{\$ 17400} \\
2014  & \$ 9202 &  \textcolor{blue}{\$ 20300} \\
2013  & \$ 9288 &  \textcolor{blue}{\$ 21400} \\ \bottomrule
\multicolumn{3}{l}{\begin{tabular}[c]{@{}l@{}}2015 compared to 2014 space systems 2019 \\ net sales in 2015 decreased \$ 97 million, or \\ 1\%, compared to 2014.
these decreases were \\ partially offset by higher net sales of  \\ approximately \$ 235 million for businesses \\ acquired in 2014; and approximately \$ 75 \\ million for the orion program due to \\ increased volume. \end{tabular}}
\end{tabular} 

& \begin{tabular}[c]{@{}l@{}}User: what was the average backlog in 2015? \\ \\Assistant: 17400 \\ \\ User: what was it in 2014? \\ \\ Assistant: 20300 \\ \\ User: what is the sum of those years? \\ \\ Assistant: 17400 + 20300 \\ \\ User: what was the average backlog in 2013? \\ \\ Assistant: 21400 \\ \\ User: what is the total sum for all 3 years? \\ \\ Assistant:  \end{tabular}  &

\begin{tabular}[c]{@{}l@{}} ChatQA-1.0-13B: \textcolor{mygreen}{17400 + 20300 + 21400} \\ \\ ChatQA-1.0-70B: \textcolor{mygreen}{17400 + 20300 + 21400}\\ \\ GPT-3.5-Turbo-0613: \textcolor{red}{9288 + 9202 + 9105} \\ \\ GPT-4-0613: \textcolor{mygreen}{17400 + 20300 + 21400} \\ \\ Ground Truth: 17400 + 20300 + 21400 \end{tabular}  \\ \midrule

\begin{tabular}{llll}
\multicolumn{4}{l}{\begin{tabular}[c]{@{}l@{}}The XVI World Rhythmic Gymnastics \\ Championships were held in Brussels, \\ Belgium, on November 20-22, 1992.\end{tabular}} \\
\toprule
\textbf{Rank} &  \textbf{Gymnast} & \textbf{Country}  &  \textbf{Point}  \\ \midrule
1  & Oxana Kostina &  Russia & 9.775 \\
2  & Maria Petrova &  Bulgaria & 9.700 \\
3  & Diana Popova &  Bulgaria & 9.625 \\ 
4  & Carmen Acedo & Spain & 9.625 \\
5  & Irina Deleanu & Romania & 9.550 \\
6  & Ekaterina S. & Ukraine & 9.525 \\
7  & Viktoriya Yani & Ukraine & 9.475 \\
\textcolor{blue}{8}  & Maria Sansaridou & Greece & 9.425 \\
\bottomrule
\end{tabular}
& 
\begin{tabular}[c]{@{}l@{}}User: What is the XVI World Rhythmic \\ Gymnastics Championships? \\ \\Assistant: The XVI World Rhythmic  \\ Gymnastics Championships were held in  \\ Brussels, Belgium, on November \\ 20-22, 1992  \\ \\ User: How many \textcolor{orange}{Gymnastics} participated \\ this championship? \\ \\ Assistant:  \end{tabular}

& 
\begin{tabular}[c]{@{}l@{}} ChatQA-1.0-13B: There are \textcolor{mygreen}{8} Gymnastics \\ participated in the championship. \\ \\ ChatQA-1.0-70B: There are \textcolor{mygreen}{8} \textcolor{orange}{gymnasts*} \\ participated in the championship.  \\ \\ GPT-3.5-Turbo-0613: There were a total \\ of \textcolor{mygreen}{8} \textcolor{orange}{gymnasts*} who participated in \\ this championship. \\ \\ GPT-4-0613: \textcolor{red}{The context does not provide} \\ \textcolor{red}{the total number of gymnasts} who \\ participated in the XVI World \\ Rhythmic Gymnastics Championships. \\ \\ Ground Truth: There are 8 gymnasts \\ participated this championship. \\ \\ \textcolor{orange}{*Both models correct the User's typo of} \\ \textcolor{orange}{``Gymnastics'' to ``gymnasts''} \end{tabular}  \\ \bottomrule

\end{tabular}
\end{adjustbox}
% \caption{Case study for ChatQA-1.0-13B, ChatQA-1.0-70B, GPT-3.5-Turbo, and GPT-4. We use \textcolor{blue}{blue} color to highlight the relevant context for the last user question. And we use \textcolor{mygreen}{green} and \textcolor{red}{red} color to highlight the correct and incorrect parts in the assistant responses.}
% \label{tab:case_study_more_1}
\end{table}

\begin{table}[]
\renewcommand{\arraystretch}{1.03}
\centering
\begin{adjustbox}{width={0.99\textwidth},totalheight={\textheight},keepaspectratio}
\begin{tabular}{lll}
\toprule
\textbf{Context}       & \textbf{Conversation}         & \textbf{Assistant Responses}     \\ \midrule
\begin{tabular}{lll}
\multicolumn{3}{l}{\begin{tabular}[c]{@{}l@{}}The 3000 metres steeplechase or 3000-meter \\ steeplechase is the most common distance for the \\ steeplechase in track and field. It is an obstacle \\ race over the distance of the 3000 metres, which \\ derives its name from the horse racing steeplechase.\end{tabular}} \\
\toprule
\textbf{Year}  &  \textbf{Athlete} & \textbf{Location} \\ \midrule
1968  & Jouko Kuha (FIN) &  Stockholm \\
1969  & Vladimiras Dudinas (URS) & Kiev \\
1970  & Kerry O'Brien (AUS) & \textcolor{blue}{Berlin} \\
1971  & Kerry O'Brien (AUS) & \textcolor{blue}{Adelaide} \\
1972  & Anders Gärderud (SWE) & Helsinki \\
1973  & Ben Jipcho (KEN) & Helsinki \\
1974  & Anders Gärderud (SWE) & Helsinki \\
1975  & Anders Gärderud (SWE) & Stockholm \\
1976  & Anders Gärderud (SWE) & Montréal \\
1977  & Michael Karst (FRG) & Stockholm \\
... \\
\bottomrule
\end{tabular}  &

\begin{tabular}[c]{@{}l@{}}User: Hello, can you tell me about the 3000 \\ metres steeplechase? \\ \\Assistant: The 3000 metres steeplechase is the \\ most common distance for the steeplechase in \\ track and field that goes over the distance of \\ the 3000 metres, which derives its name from \\ the horse racing steeplechase. \\ \\ User: Can you tell me the men that are part of it? \\ \\ Assistant: Some of the men are Henry Rono and \\ Michael Karst. \\ \\ User: Which locations did Kerry O'Brien ran for? \\ \\ Assistant: \end{tabular}  &

\begin{tabular}[c]{@{}l@{}} ChatQA-1.0-13B: He ran in \textcolor{mygreen}{Berlin, Adelaide}, \textcolor{red}{and} \\ \textcolor{red}{Stockholm}. \\ \\ ChatQA-1.0-70B: Kerry O'Brien ran in \textcolor{mygreen}{Adelaide}\\ \textcolor{mygreen}{and Berlin}. \\ \\ GPT-3.5-Turbo-0613: Kerry O'Brien ran in \textcolor{mygreen}{Berlin}, \\ \textcolor{mygreen}{Adelaide}, \textcolor{red}{and Champaign}. \\ \\ GPT-4-0613: Kerry O'Brien ran in \textcolor{mygreen}{Berlin and Adelaide}. \\ \\ Ground Truth: He ran for both Berlin and Adelaide. \end{tabular}  \\ \midrule

\begin{tabular}[c]{@{}l@{}}title: Non-EU family member traveling to Schengen \\ countries. source: You can travel visa-free only if you \\ have a ``residence'' card of a family member of a union \\ citizen'' issued under article 10 of the free movement \\ directive (or if you are a citizen of a country whose \\ nationals don't require visas in the Schengen area). A \\ residence card issued by Romania to the spouse of a \\  Romanian citizen is not necessarily an article 10 card. \\  If you do not have an article 10 card, EU countries  \\ other than Romania must issue you a visa free of charge \\ and on the basis of an expedited procedure. The grounds \\ for refusal of that visa are strictly limited. These \\ provisions are also dependent on your traveling with or \\ joining your spouse. In case you are not, you'll need a \\ regular Schengen visa. As Relaxed pointed out in a  \\ comment, it appears that your concern stems in part \\ from the fact that Romania is not yet a full participant \\  in the Schengen agreement. These rules apply to  all \\  EU countries, not just Schengen countries, because  \\ they are based on the EU right of freedom of movement. \end{tabular}

&
\begin{tabular}[c]{@{}l@{}}User: Can I travel from Romania to the Schengen \\ area visa-free? \\ \\ Assistant: You can travel visa-free only if you \\ have a ``residence card of a family member of a \\ union citizen'' issued under article 10 of the free \\ movement directive  \\ \\ User: I have a temporary residence card also, can \\ we travel together visa-free? \\ \\ Assistant: If you do not have an article 10 card, EU \\ countries other than Romania must issue you a visa \\ free of charge and on the basis of an expedited \\ procedure. \\ \\ User: How do I get a article 10 card?  \\ \\ Assistant:  \end{tabular}

&
\begin{tabular}[c]{@{}l@{}} ChatQA-1.0-13B: \textcolor{red}{The grounds for refusal of that visa} \\ \textcolor{red}{are strictly limited. These provisions are also dependent} \\ \textcolor{red}{on your traveling with or joining your spouse. In} \\ \textcolor{red}{case you are not, you'll need a regular Schengen visa.} \\ \\ ChatQA-1.0-70B: \textcolor{mygreen}{Sorry. I cannot find the answer based} \\ \textcolor{mygreen}{on the context.} \\ \\ GPT-3.5-Turbo-0613: \textcolor{red}{To obtain an article 10 card, you} \\ \textcolor{red}{would need to meet the requirements set out in the free} \\ \textcolor{red}{movement directive and apply for it through the} \\ \textcolor{red}{appropriate channels in Romania.} \\ \\ GPT-4-0613: \textcolor{mygreen}{The text does not provide information on} \\ \textcolor{mygreen}{how to obtain an article 10 card.}\textcolor{orange}{*} \\ \\ Ground Truth: Sorry. I cannot find the answer based \\ on the context. \\ \\  \textcolor{orange}{*In the evaluation, we will detect this unanswerable} \\ \textcolor{orange}{pattern and replace it with ``Sorry. I cannot find the} \\ \textcolor{orange}{answer based on the context'' to match with the} \\ \textcolor{orange}{ground truth.}  \end{tabular}  \\ \midrule

\begin{tabular}[c]{@{}l@{}}
\textcolor{blue}{title: Bengal fox. ``Vulpes bengalensis'' is a relatively}\\
\textcolor{blue}{small fox with an elongated muzzle, long, pointed} \\
\textcolor{blue}{ears, and a bushy tail about 50 to 60\% of the length} \\
\textcolor{blue}{of the head and body.} Its dorsal pelage is very variable, \\
but mostly grayish and paler ventrally; its legs tend \\
to be brownish or rufous. It is more daintily built than \\
``Vulpes vulpes''. The tail is bushy with a prominent  \\
black tip which distinguishes it from "V. vulpes". Back \\
of ears are dark brown with black margin. Its rhinarium \\
is naked and the lips are black, with small black hair \\
patches on upper part of nuzzle (shaft) in front of eyes.
\\ \\
title: Red fox. The red fox is considered a more \\
specialised form of "Vulpes" than the Afghan, corsac \\
and Bengal foxes in the direction of size and adaptation \\
to carnivory; the skull displays far fewer neotenous \\
traits than in other species ...
\\ \\
\textcolor{blue}{title: Bengal fox. The ears have the same colour as the} \\ 
\textcolor{blue}{nape or maybe darker, but not having a dark patch as in} \\
\textcolor{blue}{``V. vulpes''. Extensive variation in coat colour exists} \\
\textcolor{blue}{across populations and seasonally within populations,} \\
\textcolor{blue}{but generally varies from grey to pale brown} ...
% \\ \\
% title: Bengal fox. The Bengal fox, also known as the \\
% Indian fox, is a fox endemic to the Indian subcontinent \\ 
% from the Himalayan foothills and Terai of Nepal through \\
% southern India, and from southern and eastern Pakistan \\
% to eastern India and southeastern Bangladesh ...

\end{tabular}
&
\begin{tabular}[c]{@{}l@{}}User: what is red fox? \\ \\ 
Assistant: 
It is the largest of the true foxes and \\ one of the most widely distributed members of \\  the order Carnivora.
\\ \\ 
User: where did it originate? \\ \\ Assistant: The species is Eurasian in origin. \\ \\ 
User: how did it evolve?  \\ \\ 
Assistant: 
It is considered a more specialised form \\
of ``Vulpes'' than the Afghan, corsac and \textcolor{blue}{Bengal} \\
\textcolor{blue}{foxes} in the direction of size and adaptation \\ to carnivory. \\ \\
User: what is the appearance of \textcolor{blue}{this third type} \\ \textcolor{blue}{of fox}? \\ \\
Assistant:

\end{tabular}

& 
\begin{tabular}[c]{@{}l@{}} 
ChatQA-1.0-13B: 
\textcolor{mygreen}{The ears have the same colour as the} 
\\ 
\textcolor{mygreen}{nape or maybe darker, but not having a dark patch} \\
\textcolor{mygreen}{as in ``V. vulpes''. Extensive variation in coat colour} \\
\textcolor{mygreen}{exists across populations and seasonally within} \\
\textcolor{mygreen}{populations, but generally varies from grey to pale} \\
\textcolor{mygreen}{brown.} 
\\ \\ 
ChatQA-1.0-70B: 
\textcolor{mygreen}{``Vulpes bengalensis'' is a relatively} \\
\textcolor{mygreen}{small fox with an elongated muzzle, long, pointed} \\
\textcolor{mygreen}{ears, and a bushy tail about 50 to 60\% of the length} \\
\textcolor{mygreen}{of the head and body.} \\ \\ 
GPT-3.5-Turbo-0613: \textcolor{red}{I'm sorry, but I couldn't find} \\
\textcolor{red}{any information about a third type of fox in the} \\
\textcolor{red}{given context.} \\ \\ 
GPT-4-0613: \textcolor{red}{The context does not provide information} \\
\textcolor{red}{on a third type of fox.} \\ \\ 
Ground Truth: It is a relatively small fox with an \\
elongated muzzle, long, pointed ears, and a bushy tail \\
about 50 to 60\% of the length of the head and body.
\end{tabular}

\\ \bottomrule

\end{tabular}
\end{adjustbox}
\vspace{.1cm}
\caption{Case study for ChatQA-1.0-13B, ChatQA-1.0-70B, GPT-3.5-Turbo, and GPT-4. We use \textcolor{blue}{blue} color to highlight the relevant context for the last user question. And we use \textcolor{mygreen}{green} and \textcolor{red}{red} color to highlight the correct and incorrect parts in the assistant responses.}
\label{tab:case_study_more}
\end{table}

\newpage
\section{Human Annotated Conversational QA Data}

\subsection{Statistics of User-Agent Turns}
\begin{figure}[h]
    \centering
    \includegraphics[width=0.8\linewidth]{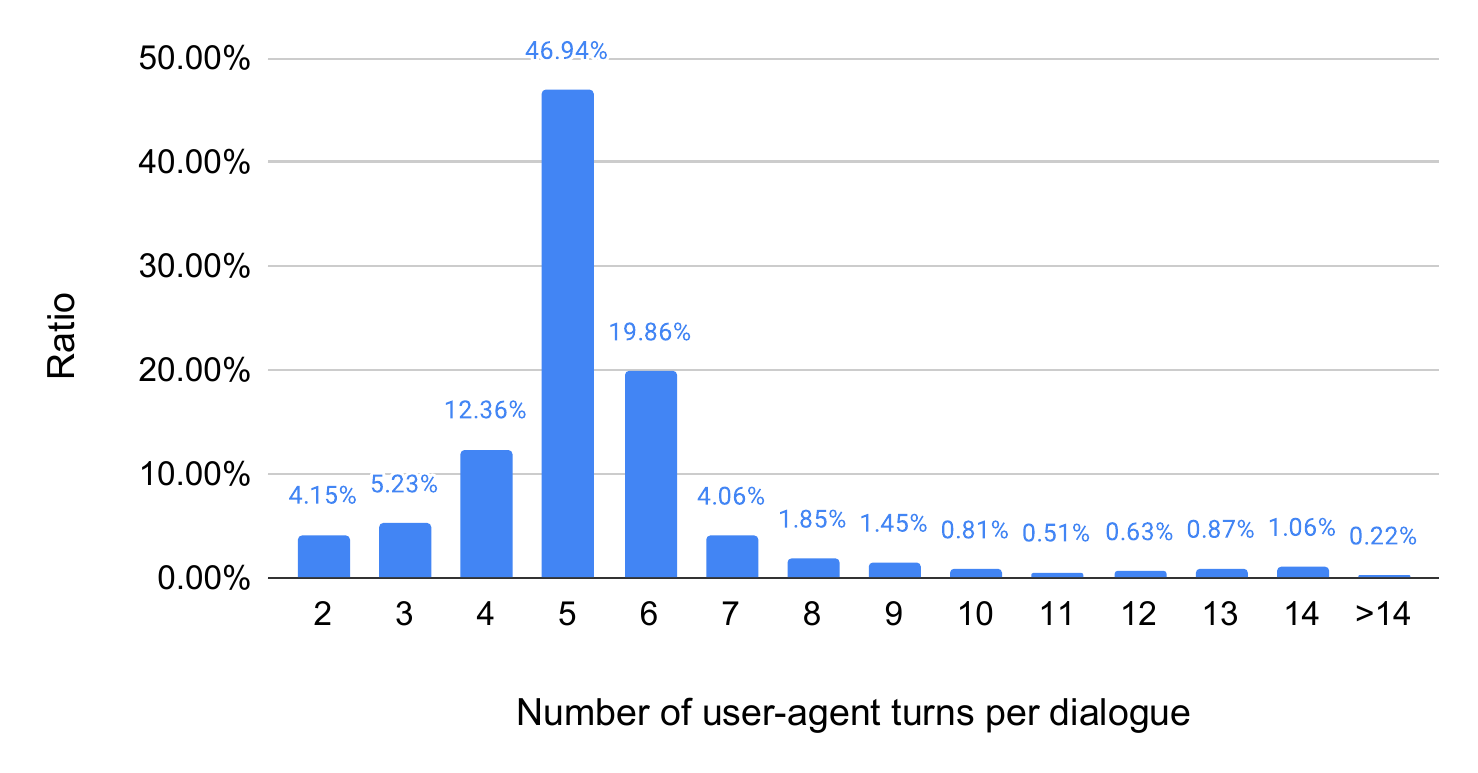}
    \caption{Statistics regarding the number of user-agent turns for each dialogue}
    \label{fig:dialog_turn_statistics}
\end{figure}

Figure~\ref{fig:dialog_turn_statistics} illustrates the statistics regarding the number of user-agent turns per dialogue over the 7k human annotated conversational QA data.

\subsection{Guidelines for Conversational QA Data Collection}
\label{appendix:data_collection}

% annotations for context location
% annotations for user-agent turns
In this section, we demonstrate the guidelines we provide for the conversational QA data collection. We request the annotators to act as both user and agent to construct the conversational QA sample based on the given document, which makes the annotation more efficient.
The guidelines consists of three parts: 1) What does conversational QA samples look like; 2) What kinds of conversational QA samples we need; and 3) What we need to annotate.

\subsubsection{What does conversational QA samples look like}
One conversational QA sample is based on a document that we provide. The user and agent behaviors are as follows:

\begin{itemize}
    \item User behavior: 1) Ask questions to agent based on the given document; 2) Answer questions from agent when agent wants to clarify something.
    \item Agent behavior: 1) Answer questions from user based on the document; 2) Ask questions to the user when the user’s question is not clear, or too general/broad.
\end{itemize}

\subsubsection{What kinds of multi-turn QA samples we need}
We list the requirements for user's questions and agent's responses below. 

\paragraph{User's Questions}
\begin{itemize}
    \item User's questions can refer to his/her previous (or multiple turns before) question.
    \item User's questions can also refer to the previous (or multiple turns before) answer from the agent.
    \item Try to use pronouns or common nouns to replace the entities mentioned before.
    \item Try to make the user's questions diverse. For the same type of questions, try to use different ways to express it in different dialogue annotations.
    \item Other user question types we need
    \begin{itemize}
        \item Given the answer from agent, ask for extra information (e.g., What else ...; any other ...; is there more ...).
        \item Switch the topic and starting a new thread in conversation.
        \item Ask two questions at the same time.
    \end{itemize}
\end{itemize}

\paragraph{Agent's Response}
\begin{itemize}
    \item Try to make the agent's answer within 1-2 sentences. If the answer has to be long, try to make it concise.
    \item Try not to directly copy the whole relevant context from the document as the answer. Instead, try to construct the answer by paraphrasing the chosen context.
    \item Try to come up with a small proportion of cases where agent asks questions to user to clarify something. Specifically, when the user's question is too broad or not clear, the agent needs to narrow down the scope by asking clarification questions to see which specific aspects that the user cares more about.
\end{itemize}

\subsubsection{What we need to annotate}
We list what we need to annotate for each conversation below.
\begin{itemize}
    \item For each document, we need to annotate the user's questions and corresponding agent's responses. The average number of user-agent turns per conversation needs to be around five.
    \item For each user's question, we need to annotate all the relevant context within the document.
\end{itemize}

\newpage
\section{Results of All Models on \textsc{ChatRAG Bench}}
\label{appendix:chatrag_bench_all}

\begin{table}[h]
\renewcommand{\arraystretch}{1.03}
\centering
\begin{adjustbox}{width={1.01\textwidth},totalheight={\textheight},keepaspectratio}
\begin{tabular}{lcccccccccccc}
\toprule
Models         & \begin{tabular}[c]{@{}c@{}}Avg. w/o\\HDial \end{tabular}    & \begin{tabular}[c]{@{}c@{}}Avg.\\All\end{tabular}  & D2D & QuAC  & QReCC & CoQA  & DoQA  & CFQA & SQA   & TCQA & HDial & INSCIT \\ \midrule
Llama2-SFT-7B* & 33.24 &  34.81 & 30.26 & 19.21 & 37.55 & 62.75 & 21.76 & 34.43 & 32.18 & 32.88 & 48.96 & 28.16 \\
Llama2-Chat-7B & 37.96  & 38.86 & 33.27 & 25.83 & 46.02 & 72.28 & 33.15 & 36.58 & 26.14 & 36.68 & 47.02 & 31.67  \\
\method-1.0-7B & 46.96 & 47.71 & 37.88 & 29.69 & 46.97 & 76.61 & 41.57 & 51.61 & 61.87 & 45.45 & 54.51 & 30.96 \\  \midrule
GPT-SFT-8B     & 32.48   & 34.46   & 31.03    & 20.07 & 37.69 & 59.24 & 21.72 & 15.44     & 40.06 & 38.17    & 52.29      & 28.86  \\
\method-1.0-8B    &  48.86   & 49.36   & 36.76    & 33.95 & 45.54 & 77.90 & 44.65 & 61.68     & 60.74 & 47.03    & 53.81      & 31.50  \\ \midrule
Llama2-SFT-13B*  & 36.08  & 37.69   & 30.68    & 21.59 & 38.25 & 69.52 & 21.70  & 41.14     & 37.85 & 35.26    & 52.22      & 28.73  \\
Llama2-Chat-13B & 39.29 & 40.34   & 34.74    & 27.89 & 47.19 & 72.50  & 32.60 & 41.54     & 25.39 & 39.25    & 49.82      & 32.52   \\
\method-1.0-13B & 50.27  & 50.86   & 38.05    & 34.28 & 48.06 & 77.23  & 43.31 & 65.44     & 66.41 & 48.88    & 56.19      & 30.79  \\ \midrule
GPT-SFT-22B*   & 37.11  &   38.83 & 32.98 & 22.83 & 39.92 & 68.19 & 22.83 & 30.54 & 47.01 & 38.07 & 54.22 & 31.66 \\
\method-1.0-22B  &  53.01   &   53.53 & 39.71 & 36.89 & 50.47 & 80.32 & 44.88 & 72.08 & 69.08 & 48.83 & 58.20 & 34.11 \\ \midrule
Llama2-SFT-70B*  &  41.85 & 43.22   & 34.42    & 25.65 & 41.88 & 73.04 & 28.21 & 46.64     & 58.90  & 37.20     & 55.52      & 30.71  \\
Llama2-Chat-70B & 44.64   & 45.21   & 36.87    & 32.47 & 49.40  & 80.41 & 38.97 & 46.85     & 37.62 & 44.31    & 50.35      & 34.88  \\ 
\method-1.0-70B & 53.89  &  54.14 & 38.90 & 41.82 & 48.05 & 78.57 & 51.94 & 73.69 & 69.14 & 50.98 & 56.44 & 31.90  \\  \midrule
Llama3-Instruct-8B & 46.76 & 46.70 & 33.59 & 33.60 & 45.70 & 75.26 & 37.33 & 58.05 & 59.72 & 44.96 & 46.20 & 32.59 \\
Llama3-\method-1.5-8B & 53.99 & 55.17 & 39.33 & 39.73 & 49.03 & 76.46 & 49.60 & 78.46 & 73.28 & 49.96 & 65.76 & 30.10 \\ \midrule
Llama3-Instruct-70B & 52.95 & 52.52 & 37.88 & 36.96 & 51.34 & 76.98 & 41.24 & 76.60 & 69.61 & 49.72 & 48.59 & 36.23  \\
Llama3-\method-1.5-70B & \textbf{57.14} & 58.25 & 41.26 & 38.82 & 51.40 & 78.44 & 50.76 & 81.88 & 83.82 & 55.63 & 68.27 & 32.31 \\ \midrule
Command R+ & 51.40 & 50.93 & 33.51 & 34.16 & 49.77 & 69.71 & 40.67 & 71.21 & 74.07 & 53.77 & 46.70 & 35.76 \\
GPT-3.5-Turbo-0613 & 50.69 & 50.37   & 34.83    & 37.17 & 50.46 & 79.33 & 41.11 & 73.15     & 60.63 & 44.30     & 47.42      & 35.27  \\
GPT-4-0613 &   54.35   & 53.90   & 34.16    & 40.29 & 52.01 & 77.42 & 43.39 & 81.28     & 79.21 & 45.09    & 49.81      & 36.34 \\
GPT-4-Turbo-2024-04-09 & \textbf{54.72} & 54.03 & 35.35 & 40.10 & 51.46 & 77.73 & 41.60 & 84.16 & 79.98 & 48.32 & 47.86 & 33.75\\
% GPT-4-Turbo-2024-04-09 \\
\bottomrule
\end{tabular}
\end{adjustbox}
\vspace{0.15cm}
\caption{Full zero-shot results on \textsc{ChatRAG Bench}. * SFT denotes that only stage-1 supervised fine-tuning is applied. We add Llama2-13B and our in-house 8B/22B models.}
\label{tab:cqa_main_full_appendix}
\end{table}

\begin{table}[h]
\renewcommand{\arraystretch}{1.03}
\centering
\begin{adjustbox}{width={1.01\textwidth},totalheight={\textheight},keepaspectratio}
\begin{tabular}{lcccccccccccc}
\toprule
Models         & \begin{tabular}[c]{@{}c@{}}Avg. w/o\\HDial \end{tabular}    & \begin{tabular}[c]{@{}c@{}}Avg.\\All\end{tabular}  & D2D & QuAC  & QReCC & CoQA  & DoQA  & CFQA & SQA   & TCQA & HDial & INSCIT \\ \midrule
Llama3-Instruct-8B & 46.76 & 46.70 & 33.59 & 33.60 & 45.70 & 75.26 & 37.33 & 58.05 & 59.72 & 44.96 & 46.20 & 32.59 \\
Llama3-Instruct-8B + Stage-2 & \textbf{54.41} & 55.59 & 39.18 & 36.54 & 49.13 & 82.43 & 48.91 & 77.99 & 72.59 & 50.88 & 66.21 & 32.04 \\ 
Llama3-\method-1.5-8B & \textbf{53.99} & 55.17 & 39.33 & 39.73 & 49.03 & 76.46 & 49.60 & 78.46 & 73.28 & 49.96 & 65.76 & 30.10 \\ \midrule
Llama3-Instruct-70B & 52.95 & 52.52 & 37.88 & 36.96 & 51.34 & 76.98 & 41.24 & 76.60 & 69.61 & 49.72 & 48.59 & 36.23  \\
Llama3-Instruct-70B + Stage-2 & \textbf{56.80} & 57.88 & 40.22 & 40.58 & 50.66 & 76.44 & 51.48 & 82.75 & 80.30 & 56.21 & 67.57 & 32.58 \\
Llama3-\method-1.5-70B & \textbf{57.14} & 58.25 & 41.26 & 38.82 & 51.40 & 78.44 & 50.76 & 81.88 & 83.82 & 55.63 & 68.27 & 32.31 \\ \bottomrule
\end{tabular}
\end{adjustbox}
\vspace{0.15cm}
\caption{Studies on directly applying stage-2 instruction tuning to the powerful Llama3-Instruct models.}
\label{tab:llama3_instruct_plus_stage2}
\end{table}

In Table~\ref{tab:cqa_main_full_appendix}, we report zero-shot results of all ChatQA models on \textsc{ChatRAG Bench}. ChatQA models generally give great improvements compared to the corresponding SFT/Chat/Instruct counterparts. 

In Table~\ref{tab:llama3_instruct_plus_stage2}, we study the effectiveness of directly applying stage-2 instruction tuning to the powerful Llama3-Instruct models. Llama3-Instruct models can be considered as those that have already finished the Stage-1 SFT training. They use a much larger dataset than the stage-1 training for Llama3-ChatQA-1.5. However, we find that ``Llama3-Instruct-8B/70B + Stage-2'' performs comparable to Llama3-ChatQA-1.5-8B/70B. This further suggests that the stage-2 instruction tuning plays a key role in improving model's capability in conversational QA and RAG.

\end{document}